\documentclass[conference]{IEEEtran}
\IEEEoverridecommandlockouts
\usepackage{cite}
\usepackage{amsmath,amssymb,amsfonts}
\usepackage{algorithmic}
\usepackage{subcaption}
\usepackage{graphicx}
\usepackage{textcomp}
\usepackage{xcolor}
\usepackage{booktabs,floatrow}
\usepackage{comment}
\usepackage{amsmath}
\usepackage{url}
\usepackage{makecell}
\usepackage{svg}
\usepackage{multicol}

\def\BibTeX{{\rm B\kern-.05em{\sc i\kern-.025em b}\kern-.08em
    T\kern-.1667em\lower.7ex\hbox{E}\kern-.125emX}}
\begin{document}

\title{Learning Pedestrian Actions to Ensure Safe Autonomous Driving\\
}

\author{\IEEEauthorblockN{Jia Huang}
\IEEEauthorblockA{\textit{Dept of Mechanical Engineering} \\
\textit{Texas A\&M University}\\
College Station, Texas, USA \\
jia.huang@tamu.edu}
\and
\IEEEauthorblockN{Alvika Gautam}
\IEEEauthorblockA{\textit{Dept of Mechanical Engineering} \\
\textit{Texas A\&M University}\\
College Station, Texas, USA \\
alvikag@tamu.edu}
\and
\IEEEauthorblockN{Srikanth Saripalli}
\IEEEauthorblockA{\textit{Dept of Mechanical Engineering} \\
\textit{Texas A\&M University}\\
College Station, Texas, USA \\
ssaripalli@tamu.edu}

}

\maketitle

\begin{abstract}
%\alvika{still working on this}
To ensure safe autonomous driving in urban environments with complex vehicle-pedestrian interactions, it is critical for Autonomous Vehicles (AVs) to have the ability to predict pedestrians' short-term and immediate actions in real-time. In recent years, various methods have been developed to study estimating pedestrian behaviors for autonomous driving scenarios, but there is a lack of clear definitions for pedestrian behaviors. In this work, the literature gaps are investigated and a taxonomy is presented for pedestrian behavior characterization. Further, a novel multi-task sequence to sequence Transformer encoders-decoders (TF-ed) architecture is proposed for pedestrian action and trajectory prediction using only ego vehicle camera observations as inputs. The proposed approach is compared against an existing LSTM encoders decoders (LSTM-ed) architecture for action and trajectory prediction. The performance of both models is evaluated on the publicly available Joint Attention Autonomous Driving (JAAD) dataset, CARLA simulation data as well as real-time self-driving shuttle data collected on university campus. Evaluation results illustrate that the proposed method reaches an accuracy of 81\% on action prediction task on JAAD testing data and outperforms the LSTM-ed by 7.4\%, while LSTM counterpart performs much better on trajectory prediction task for a prediction sequence length of $25$ frames.
\end{abstract}


\section{Introduction and Related Work} \label{sec:intro}
Safe interaction with other road users is a crucial challenge for the development and implementation of intelligent vehicles to drive completely autonomously. Urban driving settings are more prone to pedestrian accidents \cite{crash_report}, thus anticipation and estimation of pedestrian behavior on the streets and taking timely maneuvering decisions is one of the key requirements for the adoption of AVs in mixed traffic, such as urban driving scenarios or autonomous shuttles for last mile connectivity in unstructured driving settings like university campuses, small towns, etc \cite{urban_survey,urban_survey2}.

A significant body of literature is focused on pedestrian modeling through purely vision-based pedestrian detection and tracking \cite{deepsort,wang2018repulsion,zhang2018occlusion}, or trajectory prediction \cite{xu2018encoding,korbmacher2022review,shi2022social}. Although these methods show promising results, it is not enough for pedestrian-aware autonomous driving as pedestrian behavior patterns are highly dynamic (with sudden changes in motion plans and direction \cite{ferguson2015real,schneider2013pedestrian}) and sensitive to even small environmental changes. 

As a higher level behavior modeling, one of the most critical and commonly studied problems is anticipating or predicting pedestrian's behavior  to \textit{cross/not cross} the vehicle/street. This is typically referred to in the literature as pedestrian \emph{intention}. Several works explore intention prediction using measurable or observable entities such as trajectories from visual data \cite{bouhsain2020pedestrian_LSTMPaper}, pedestrians' actions, skeletal postures \cite{HMM,pedestrian_pose} and spatiotemporal cues along with pose data \cite{spatiotemporal}. Action classification based on future scene prediction \cite{gujjar2019classifying} and predicting intent using Markov property \cite{HMM} are also explored where the hidden states are pedestrian intents. Machine learning approaches are also used to model intent as a binary classification problem where the output is crossing/not crossing \cite{bouhsain2020pedestrian_LSTMPaper,spatiotemporal} and these approaches are validated on state-of-the-art pedestrian behavior datasets \cite{JAAD_cite1,JAAD_cite2,PIE_dataset,psi_dataset}. 

In this work, we briefly discuss the gaps in the literature with respect to definitions of pedestrian behaviors and present a taxonomy to characterize various levels of behavior relevant to autonomous driving scenarios. We propose a multi-task sequence to sequence Transformer encoders-decoders architecture for pedestrians' trajectory and action prediction, and compare against the state of the art PV-LSTM model proposed in \cite{bouhsain2020pedestrian_LSTMPaper} using the same inputs and tasks. We evaluate the efficacy of our approach on simulation (CARLA \cite{CARLA}), JAAD \cite{JAAD_cite1,JAAD_cite2} dataset using ground truth information. Finally, we collect camera data on a self-driving shuttle on a university campus, and perform an end-to-end evaluation on both models using vision-based pedestrian tracking information as inputs. YOLOv5 \cite{yolov5} detector and DeepSORT \cite{deepsort} are used for pedestrian detection and tracking respectively.

\section{Taxonomy}
Despite significant work within literature, there is a lack of consistent definitions and taxonomy for pedestrian behavior. Further, existing works use action recognition, trajectory estimation, pose based motion prediction as surrogates to the actual pedestrian intention. 
\begin{figure}[h]
\centering
\includegraphics[width=0.9\textwidth,trim=2cm 0cm 1cm 1cm]{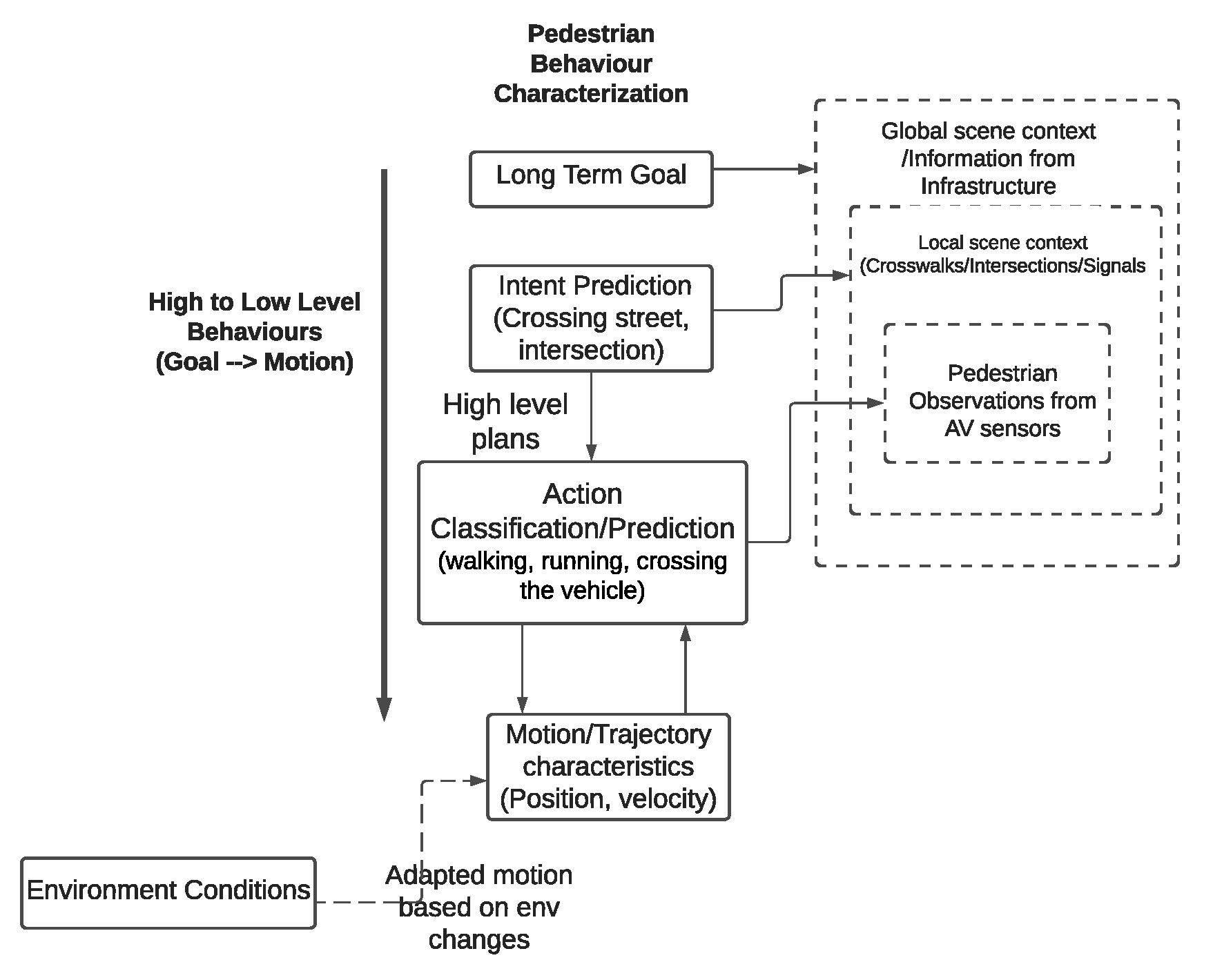}
\caption{Classification of levels of pedestrian behaviors}
\label{fig:behaviour}
\end{figure}
To this end, we propose and discuss a taxonomy of pedestrian behavior characterization in this work. A high-level flowchart of this taxonomy is shown in Fig. \ref{fig:behaviour}. In a traffic scenario, pedestrians can have a long-term objective or destination,  short-term objective/actions (e.g. crossing an intersection or street), and immediate actions (e.g. crossing near a vehicle on the street). Typically, behavior characterizations include answers to questions such as what exactly a pedestrian plans to do or what their objective (\textit{Intent}) is, what exactly a pedestrian is about to do (\textit{Action}), what kind of trajectory the pedestrian is going to take (\textit{Motion}). There exists a hierarchical causal relationship between these, where pedestrians take actions based on their high-level plans or intent and the motion characteristics of these actions can change in response to traffic/road situations. The requirement for environment/traffic scene information increases proportionally with the required level of behavior estimation. According to our taxonomy, true intent prediction (crossing/not crossing a street) requires local scene context. Without this information, it is ambiguous to differentiate between a pedestrian's \emph{intent} to cross a street versus a pedestrian's \emph{action} of crossing in front of a vehicle. 

Because pedestrian behavior depends on a number of factors like  traffic, environment, and individual pedestrian characteristics, we propose that behavior estimations should explicitly assert the span for which these estimations apply. For instance, consider a pedestrian waiting at a bus stop, if they step on the road to check for the bus, this might be interpreted as a crossing action by an approaching ego vehicle following a trajectory-based approach, but this is not reflective of the pedestrian's overall objective or intent. Thus, we define actions with their motion characteristics as low-level behaviors with respect to an ego vehicle, which can be characterized to some extent solely by pedestrian observations and are critical to safe path planning and navigation of AVs. Intent is a high-level behavior characterization that is not directly observable and cannot be estimated by simply using the pedestrian's trajectory but requires a deeper inference from past behavior, past actions, contextual scene information, etc.

In this paper, we focus on low-level behavior characterization of a pedestrian (crossing/not crossing) with respect to a single vehicle, i.e. action prediction relying purely on visual observations of the ego vehicle camera without characterizing it as intent because contextual scene information and inference from past actions is neither known apriori nor estimated.
% We develop a multi-task sequence to sequence Transformer Encoder-Decoder architecture for pedestrians' trajectory and action prediction and we use the LSTM model proposed in \cite{bouhsain2020pedestrian_LSTMPaper} for performance comparison. We evaluate the efficacy of our approach on simulation (CARLA), state of the art JAAD dataset using ground truth information. Finally, we collect onboard camera data on a self-driving shuttle in the university campus, and an end to end evaluation of both the models is performed using vision based pedestrian tracking information as input to the trajectory-action prediction models. Yolov5 detector and DeepSort are used for pedestrian detection and tracking respectively.

\section{Methodology}\label{sec:method}

\subsection{Problem Formulation}\label{subsec:probform}
We formulate pedestrian action and trajectory prediction as a multi-objective learning problem.
Let the pedestrian's position and speed at time $t$ be denoted as $p_{t}$ and $s_{t}$ respectively. Let the pedestrian's action at time t be denoted by,
\begin{align}
%\[
    A_{t} = \left\{
        \begin{array}{ll}
            \mbox{\tt 0} & \mbox{if not crossing} \\
            \mbox{\tt 1} & \mbox{if crossing}
        \end{array}
    \right.,
%\]
\end{align}
At time $t$, given the historical position and speed trajectories of length $m+1$ denoted as,
\begin{align}
  P_{t} = \{ p_{t},p_{t-1}, p_{t-2}...p_{t-m} \}, \label{eq:pos_seq} \\
  S_{t} = \{ s_{t},s_{t-1}, s_{t-2}...s_{t-m} \}, \label{eq:speed_seq}
\end{align}
we learn the probability distribution,
\begin{align}
     Pr ( \hat{P}^{t+n}, \hat{A}^{t+n} | P_{t}, S_{t} ),
 \end{align}
where, $\hat{P}^{t+n}$, $\hat{A}^{t+n}$ are predicted next $n$ positions and actions.

Next, we discuss the high-level architectural details of the proposed TF-ed and the PV-LSTM models, including model inputs and outputs followed by an individual description of both architectures.

\subsection{Architecture}\label{subsec:Arch_LSTM_TF}
In this work, we propose the multi-task Transformer encoders-decoders(TF-ed) architecture and compare its performance with the LSTM encoders-decoders (LSTM-ed) architecture of PV-LSTM model in \cite{bouhsain2020pedestrian_LSTMPaper}. The high-level diagram of the overall TF-ed framework is shown in Fig. \ref{fig:model architecture}. For both LSTM-ed and TF-ed, sequences of observed speeds and positions of pedestrians are encoded through the corresponding encoders. The high-dimensional hidden features are concatenated at the output of the two encoders and passed to the speed and the action decoder. Finally, the outputs of the decoders are back-projected to sequences of predicted speeds and actions with a dimension of 4 and 2, respectively.
\subsubsection{Model input and output}
Spatial bounding box (bbox) coordinates around pedestrians are used to encode pedestrians' locations. The bbox coordinates are typically available either in the form of ground truth from state-of-the-art trajectory data sets or can be obtained as outputs of a real-time multi-object tracking algorithm. Additionally, pedestrians' speed is another feature input that is obtained by subtracting the bbox coordinates of two consecutive frames for the same pedestrian. Formally, for a pedestrian \(i\), given a sequence of historical position and speed observations (Eqns. \eqref{eq:pos_seq} and \eqref{eq:speed_seq}), where ${p}_t^{(i)}$ is the bbox center coordinates (x, y, width, height) and ${s}_t^{(i)}$ is the corresponding speed,
\begin{align}
 \textbf{p}_t^{(i)} = (x_t^{(i)}, y_t^{(i)},w_t^{(i)},h_t^{(i)}),\\
 \textbf{s}_t^{(i)} = (\Delta x_t^{(i)}, \Delta y_t^{(i)},\Delta w_t^{(i)},\Delta h_t^{(i)}),
\end{align}
we predict a sequence of pedestrians' speeds and future actions for time instance $t+1$ to $t+n$, the future positions can be computed from predicted speeds.
% \begin{align}
%  L_{obs}=\{\textbf{l}_t^{(i)}\}_{t=t-m+1}^t,\\
%  S_{obs}=\{\textbf{s}_t^{(i)}\}_{t=t-m+1}^t
% \end{align}\(L_{obs}=\{\textbf{l}_t^{(i)}\}_{t=t-m+1}^t\) and \(S_{obs}=\{\textbf{s}_t^{(i)}\}_{t=t-m+1}^t\) of m observations up to time t where 

\subsubsection{Transformer encoders-decoders (TF-ed)}
\begin{figure}
\centering
\begin{subfigure}{0.85\textwidth}
    \includegraphics[width=\textwidth]{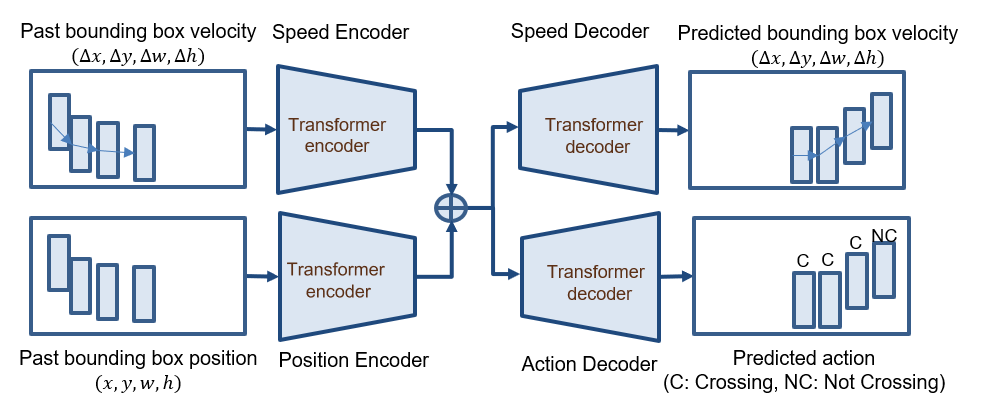}
        \caption{High level diagram of Transformer encoders-decoders (TF-ed) architecture}
        \label{fig:model architecture}
\end{subfigure}
\hfill
\begin{subfigure}{0.95\textwidth}
    \includegraphics[width=\textwidth]{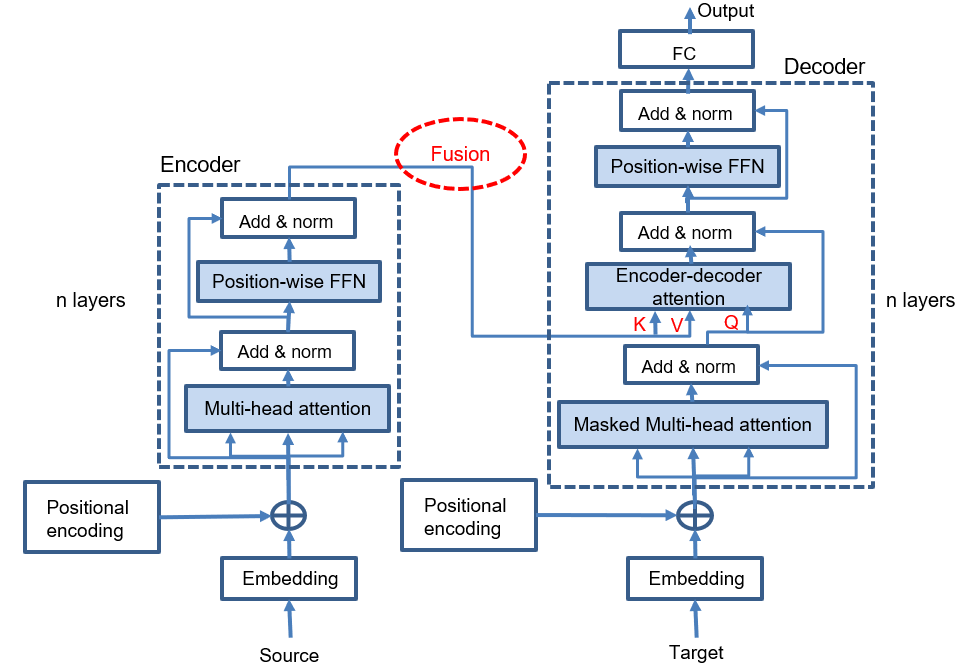}
        \caption{The Transformer architecture.The red dash line circle shows the part where the fusion of two encoders' output takes place}         
        \label{fig:Transformer} 
\end{subfigure}
\caption{Schematic diagram for the overall TF-ed framework}
\label{fig:model}
\end{figure}

TF network was first proposed in \cite{attention} for Natural Language Processing. Unlike LSTM which processes the input sequence step by step, TF processes the whole embedding sequence at once, which enables parallel training, see Fig. \ref{fig:Transformer}. TF consists of an encoder and a decoder. The source and target sequence embeddings are added with positional encoding before being fed into the encoder and the decoder that stack modules based on self-attention. 
\paragraph{Input Embedding} The speed and position source and target inputs are first embedded onto a higher D-dimensional space by a fully connected layer.
\paragraph{Positional Encoding}Each input embedding is time-stamped by adding a positional encoding mask PE of the same time, PE is calculated as,

\begin{equation}
  PE_{t,d}=\begin{cases}
    sin(\frac{t}{10000^{\frac{d}{D}}}), & \text{if d is even}.\\
    cos(\frac{t}{10000^{\frac{d-1}{D}}}), & \text{if d is odd}.
  \end{cases}
\end{equation}

where $d$ denotes the $d^{th}$ dimension of the overall $D$-dimensional embedding. 
\paragraph{Multi-Head Self-Attention} The capability of the network to capture sequence non-linearities lies mainly in the attention models. The inputs are embedded into Queries (Q), Keys (K), and Values (V) vectors. Q and K are used to calculate the attention matrix which reflects the correlation among the input sequence through a scaled dot product and a softmax layer. The attention matrix is then used to weight V to make sure every part of the output is merged with the temporal information from other parts in the sequence as follows, 
\begin{align}
    Attention(Q,K,V) = softmax\Bigl(\frac{QK^T}{\sqrt{d_k}}\Bigr)V
\end{align}
Specifically, in the encoder-decoder attention layer, Q is from the outputs of the previous decoder layer, K and V are from the encoder outputs. In the decoder self-attention, Q, K and V are all from the outputs of the previous decoder layer. However, each position in the decoder is allowed to only attend to all positions in the decoder up to that position, ensuring that the prediction depends only on already generated output predictions. During evaluation, since there are no ground truth targets unlike training using ground truth data, previously predicted output of the decoder is used as the new target input iteratively. The overall architecture of our proposed TF-ed is presented in Fig. \ref{fig:model architecture}. Two TF encoders process the speed and position inputs in parallel, the K and V outputs from both encoders are fused through concatenation and then passed to separate decoders for speed prediction and action prediction. The predicted speed outputs are extended as an output sequence which is used to compute the future trajectories of the pedestrians. 

\subsubsection{LSTM encoders-decoders (LSTM-ed)}
\begin{figure}[!h]
\centering
\includegraphics[scale=0.35]{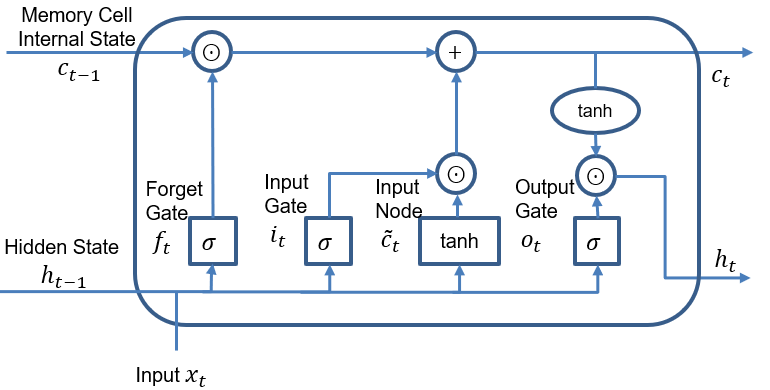}
\caption{Computing the hidden state in an LSTM model}
\label{fig: LSTM}
\end{figure}

LSTM is capable of learning long-term dependencies from time-series data. In addition to having characteristics of recurrent neural networks, LSTM also includes a purpose-built memory cell $c_t$ which consists of an input gate $i_t$, a forgot gate $f_t$, and an output gate $o_t$, see Fig. \ref{fig: LSTM}. These gates control and protect the information flow in each memory cell.
Given an input vector,
\begin{align}
x = (x_1, x_2,...,x_T),
\end{align}
the hidden layer output $h_t$ of each LSTM memory cell is calculated and updated at each time step \(t\) from $t=1$ to $t=T$ according to the following set of equations: 
\begin{align}
  i_t = \sigma(W_{xi}x_t+W_{hi}h_{t-1}+b_i), \\
  f_t = \sigma(W_{xf}x_t+W_{hf}h_{t-1}+b_f), \\
  o_t = \sigma(W_{xo}x_t+W_{ho}h_{t-1}+b_o), \\
  \tilde{c}_t = tanh(W_{xc}x_t+W_{hc}h_{t-1}+b_c),\\
  c_t = f_t \odot c_{t-1}+i_t \odot \tilde{c}_t, \\
  h_t = o_t \odot tanh(c_t),
\end{align}
where $W_*$ and $b_*$ are the weight matrices and biases for the three gates and memory cell, and $\sigma(x)$ is the sigmoid function,
\begin{align}
  \sigma(x)=\frac{1}{1+e^{-x}}
\end{align}

The overall LSTM-ed architecture used for comparison is adopted from paper\cite{bouhsain2020pedestrian_LSTMPaper}. Given observed input sequences of speed and position, the speed encoder \(LSTM_s^{enc}\) and position encoder \(LSTM_p^{enc}\) generate corresponding speed and position hidden state at time step t, which are concatenated to a single hidden state containing all features and then passed to the decoders for trajectory and action prediction task separately.

\section{Experimental Evaluation}\label{sec:expEval}
We evaluate the methodology described in Section \ref{sec:method}, both LSTM-ed and TF-ed architectures across three scenarios for action and trajectory prediction: ~(1) Simulation scenarios in CARLA simulator ~(2) State of the art JAAD dataset ~(3) Data collected on a shuttle in a university campus driving scenario.
Through CARLA and JAAD datasets, we aim to evaluate the performance of action and trajectory prediction individually and thus use ground truth (GT) data as input for both models. We then evaluate the same on the data collected on a self-driving shuttle, where pedestrian data obtained through a vision-based tracking algorithm is used as input to the models. The end-to-end evaluation (image input to action and trajectory prediction) is essential to verify the efficacy of the approach for real-world navigation solutions of AVs in the presence of noisy conditions and potential tracking errors/failures.

\subsection{Evaluation Setup} \label{subsec:eval_scenarios}

\begin{figure}[!h]
\centering
\begin{subfigure}{0.45\textwidth}
    \includegraphics[width=\textwidth]{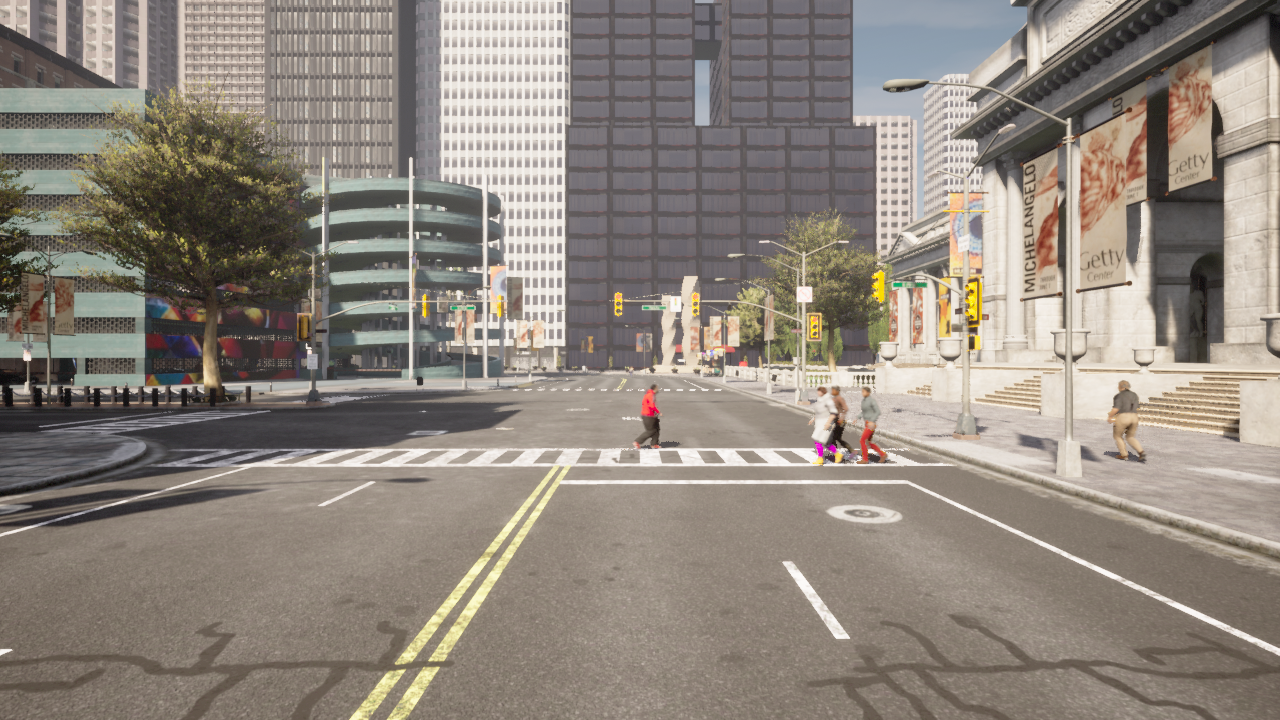}
        \caption{Ego vehicle camera view}
        \label{fig:carla_front}
\end{subfigure}
\hfill
\begin{subfigure}{0.45\textwidth}
    \includegraphics[width=\textwidth]{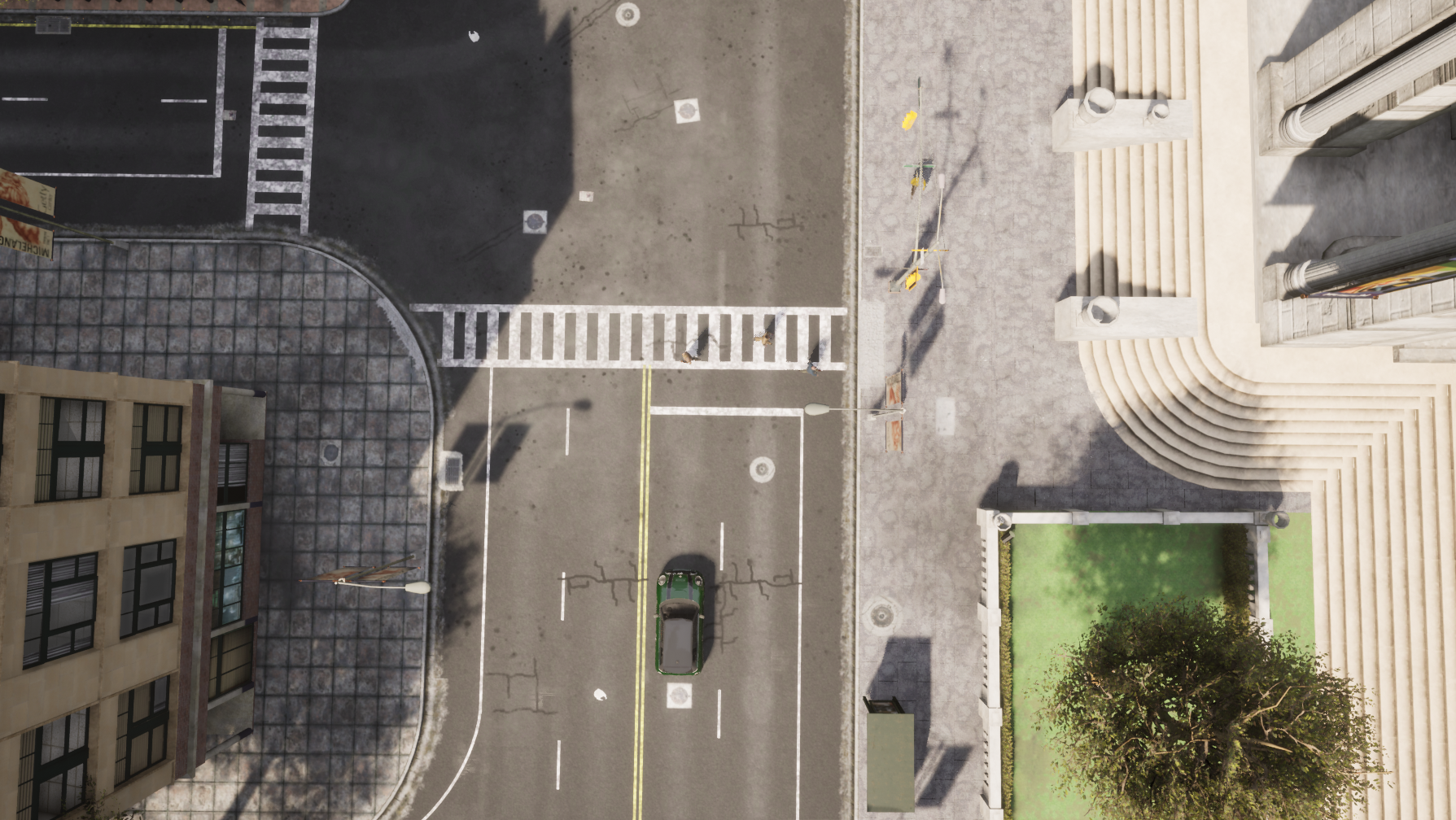}
        \caption{Top down view of the map}
        \label{fig:carla_topdown2}
\end{subfigure}
\caption{CARLA simulation platform a) CARLA vehicle camera view with pedestrian
in frame b) CARLA overall top-down view }
\label{fig:carla setup}
\end{figure}

\paragraph{CARLA}
CARLA \cite{CARLA} is an open-source simulator for autonomous driving research. In our work, an urban driving environment in CARLA (Town10HD) is used as the simulation environment. Fig. \ref{fig:carla_topdown2} shows the bird's eye view of this environment. An ego-vehicle with an on-board camera with a resolution of 1920x1080 and a FOV of 90$^{\circ}$ is spawned close to the crosswalk (camera view shown in Fig. \ref{fig:carla_front}), a walker and its controller are spawned together on the right sidewalk. The walker navigates along the sidewalk and stops close to the crosswalk for a moment and starts to cross until he reaches the other side of the road. During the simulation, the GT bboxes and corresponding action labels like crossing/not crossing were programmed to be generated automatically. The whole sequence contains 145 frames with GT bboxes and action annotations for pedestrians.

\paragraph{Joint Attention Autonomous Driving (JAAD) Dataset}:
JAAD \cite{JAAD_cite1,JAAD_cite2} dataset focuses on joint attention in the context of autonomous driving in urban city driving settings. The dataset is designed to study pedestrian and driver behaviors such as when pedestrians cross/not cross. JAAD contains videos captured with a front-view camera at 30 FPS with a resolution of 1920x1080 pixels under various naturalistic scenes, and light conditions. There are 346 videos with 82032 frames, where the length of videos ranges anywhere from 60 frames to 930 frames. 2793 pedestrians are annotated with bboxes,  686 out of which have behavioral tags including actions like walking, standing, crossing, looking (at the traffic), etc. In the dataset, \emph{walking} behavior annotation includes sub-behaviors: ``walking while crossing'' or ``walking but not crossing''. For this evaluation, we only use tags associated with \emph{crossing} behavior which is a subset of walking for action prediction so as to maintain the same data processing as that in \cite{bouhsain2020pedestrian_LSTMPaper} for consistency in comparison.
% To clarify, walking consists of walking but not crossing and walking while crossing, namely crossing tags are subsets of walking tags. Here, we only focus on crossing tags for action prediction.

\begin{figure*}[!h]
\centering
\hspace{-0.2in}
\begin{subfigure}{0.24\textwidth}
    \includegraphics[width=\textwidth]{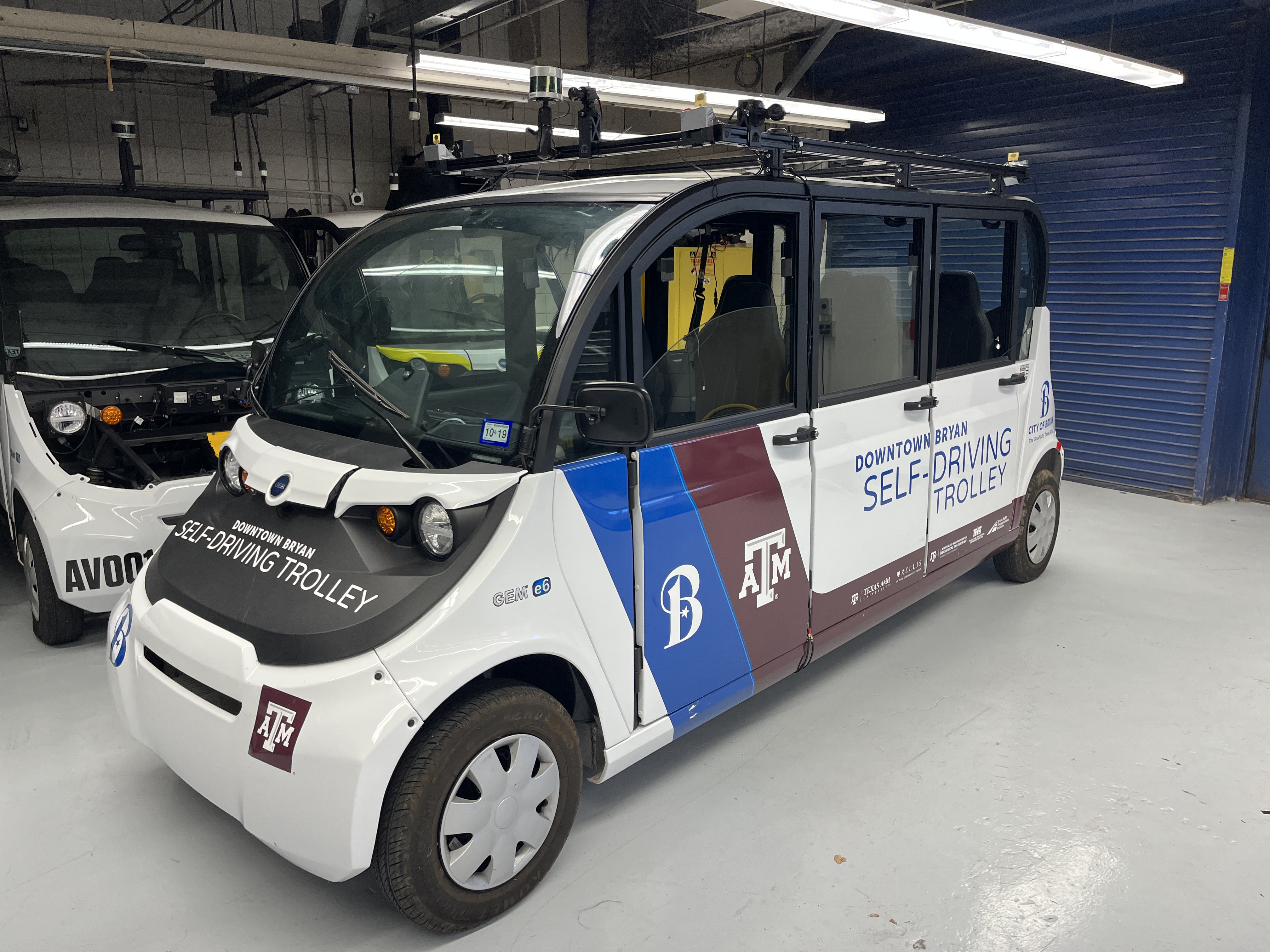}
        \caption{Shuttle vehicle setup for data collection in campus}
        \label{fig:golf_setup}
\end{subfigure}
\hfill
\begin{subfigure}{0.33\textwidth}
    \includegraphics[width=\textwidth]{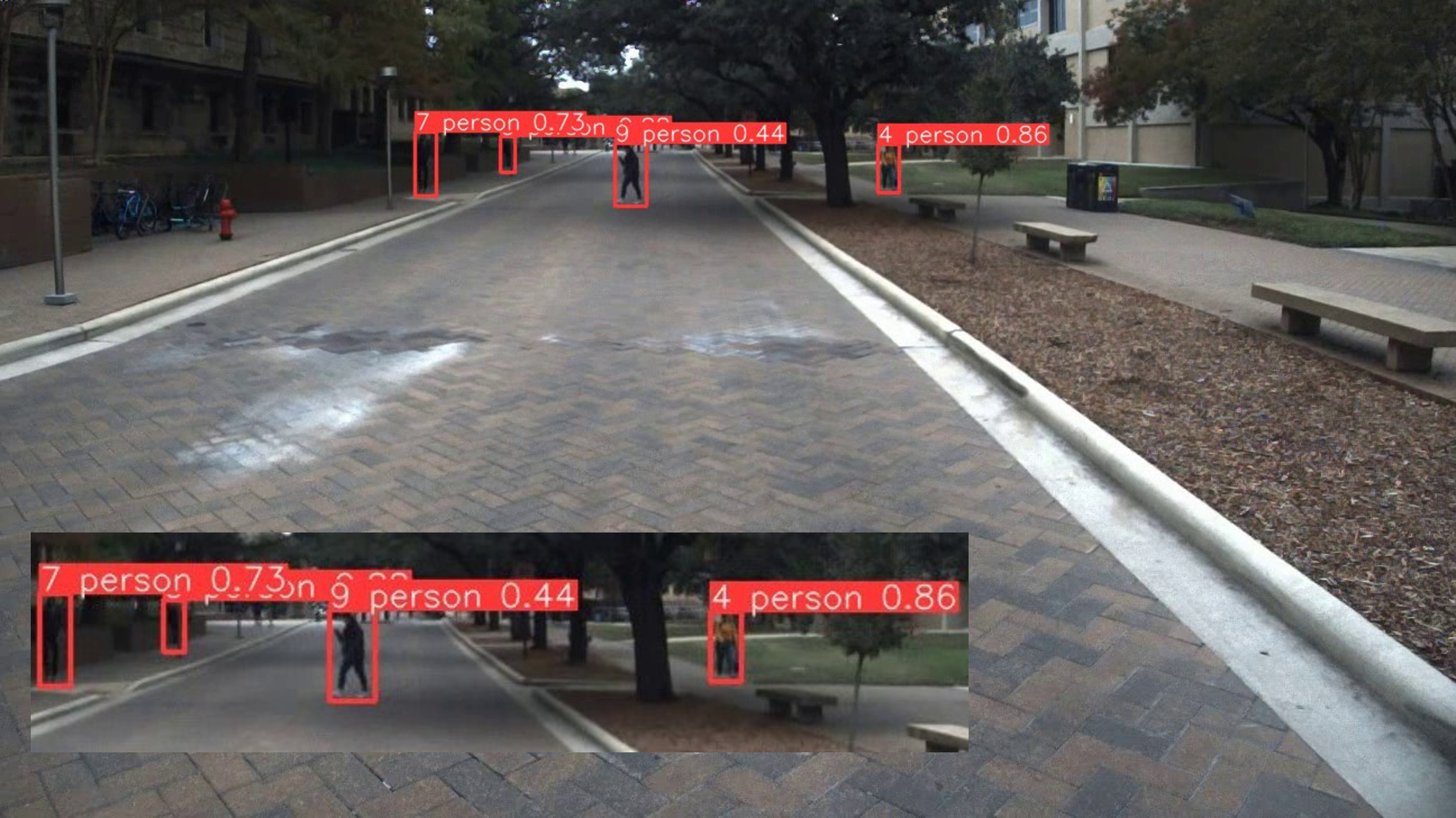}
        \caption{Case 1: All pedestrians are detected successfully}
        \label{fig:Tracking success}
\end{subfigure}
\hfill
\begin{subfigure}{0.33\textwidth}
    \includegraphics[width=\textwidth]{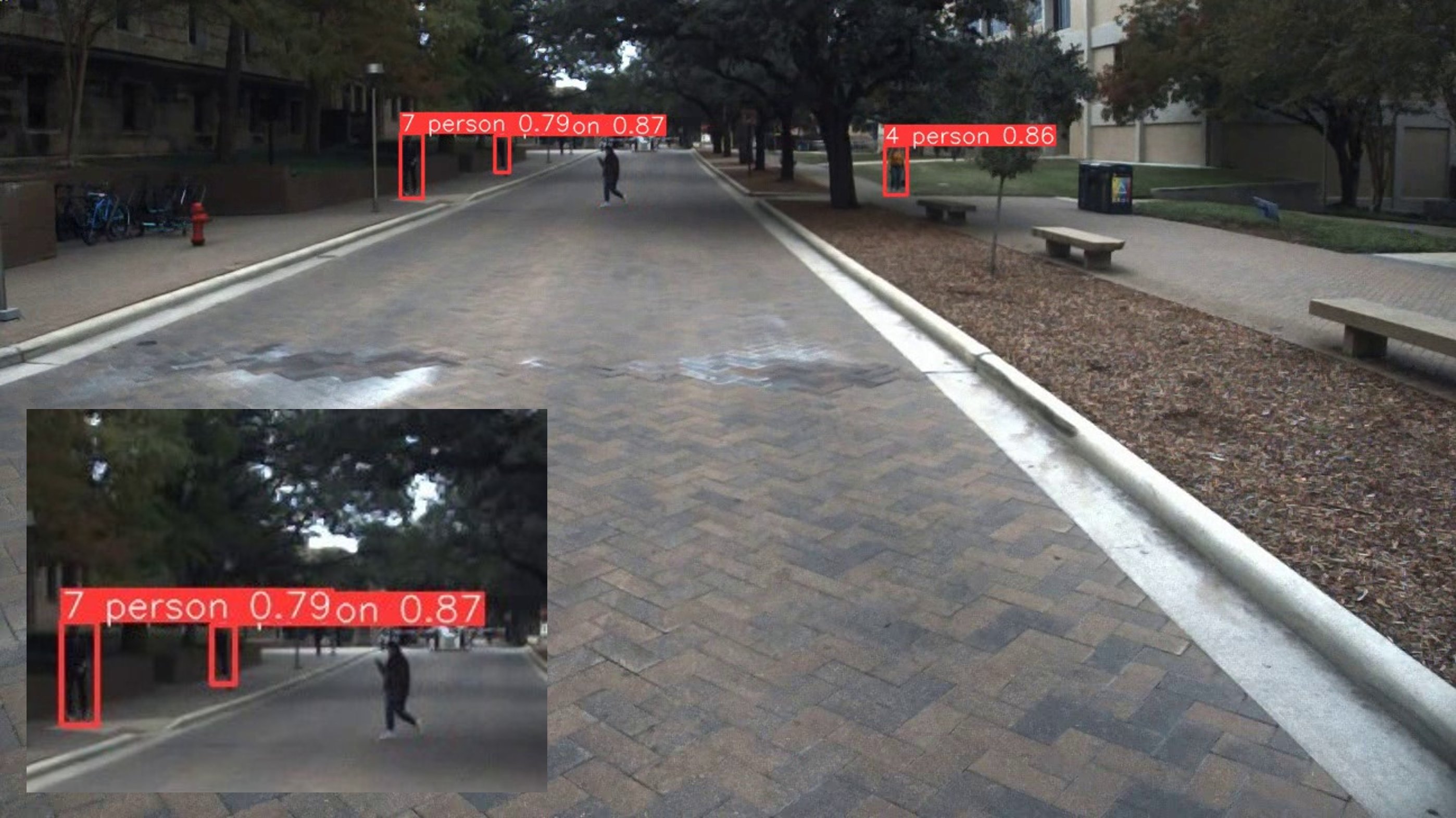}
        \caption{Case 2: One of the pedestrians is not detected}
        \label{fig:Tracking failure}
\end{subfigure}
\caption{Examples of pedestrian detection and tracking results a) represents the successful case when all pedestrians are detected and b) represents a failure case when the pedestrian in the middle of the road is not detected. The inset figures show a closer view of both cases}
\label{tracking}
\end{figure*}
\paragraph{Real-time data}
The vehicle used for real-time data collection is a Polaris GEM e6 purchased from Autonomous Stuff, shown in Fig. \ref{fig:golf_setup}. The vehicle is equipped with GPS, LIDAR, and a camera mounted on the roof of the vehicle in the front center. All control and sensing software is integrated using the Robot Operating System (ROS) \cite{ROS} and the data was collected at a frame rate of 15 FPS on the Texas A\&M university campus. The shuttle was driven manually at an average speed of $10$ miles per hour. The data consists of 298 frames with a resolution of 1296x728 pixels, covering typical pedestrian crossing/not crossing scenarios on campus. The data was intentionally collected in less-than-ideal lighting conditions in the evening for realistic testing.

\subsection{Implementation details}\label{subsec:imple}
\subsubsection{Data Pre-processing}
Model inputs for action prediction are in the form of bbox pixel coordinates $(x,y,w,h)$ of the pedestrian. For JAAD and CARLA evaluations, the ground truth of the bbox coordinates is used. Further, the pedestrian ID and behavioral tags are extracted from data annotations.
Once the whole sequence of the bbox coordinates and corresponding action class are obtained for each pedestrian throughout all the frames, the sequence is further divided into sub-sequences, each containing observations sequence with length $O$ and corresponding ground truth sequence with length $T$ iteratively for the dataloader.

\begin{figure}[!h]
\centering
\includegraphics[width=\textwidth]{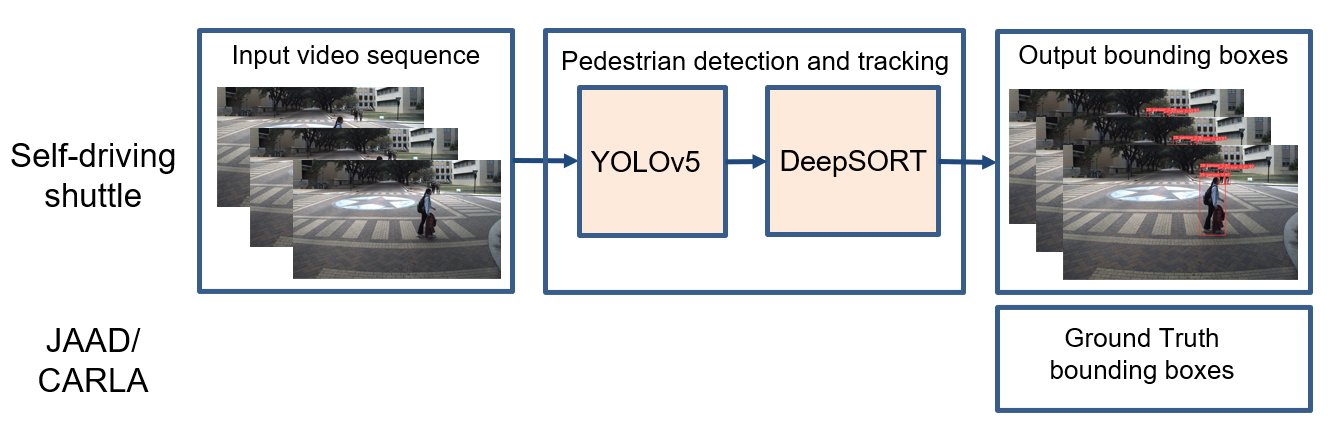}
\caption{Data input to the model. For JAAD and CARLA data, ground truth bboxes are provided directly as data input, while for self-driving shuttle data, the framework is evaluated end-to-end, and real time pedestrian tracking output is used as input for action prediction.}
\label{fig:data input}
\end{figure}

For self-driving shuttle data evaluation, a vision-based detection/tracking algorithm is integrated into the pipeline and the tracking output is used as an input to action prediction models as shown in Fig. \ref{fig:data input}. YOLOv5 \cite{yolov5} and DeepSORT \cite{deepsort} algorithms are used to detect and track multiple pedestrians in video sequences.  Since only the pedestrians' bbox information is required in the study, the object class to be detected is set to 0 (person) when running inferences based on the weight \emph{crowdhuman-yolov5m} pre-trained on the MS COCO \cite{mscoco} dataset for YOLOv5 and $osnet\_x1\_0$ for DeepSORT.

\subsubsection{Training and Testing}
Both LSTM-ed and TF-ed models were trained on JAAD training and validation sets, then evaluated on CARLA simulation, JAAD testing sets, and self-driving shuttle data respectively. However, the evaluation of self-driving shuttle data can only provide qualitative results since there is no ground truth annotation. JAAD dataset was split into training, validation, and testing with a ratio of 0.7:0.1:0.2.
\begin{table}[!h]
\noindent
\small
%\qquad\qquad
\subfloat[Training Parameters and Specifications]{
%\resizebox{\columnwidth}{!}{
\begin{tabular}{|p{2.8cm}|p{4.3cm}|} \hline
Batch size & \makecell[l]{JAAD: 128 \\ CARLA, shuttle : 1} \\ \hline
Optimizer & Adam \\ \hline
Initial Learning Rate & 0.0001 \\ \hline
Loss Function & \makecell[l]{Action Prediction : Binary CE \\ Trajectory Prediction : MSE} \\ \hline
Training Epochs & 100 \\ \hline
GPU & Nvidia GeForce GTX 1080 Ti \\ \hline
\end{tabular}}
\qquad\qquad
\subfloat[TF-ed model parameters]{
\begin{tabular}{|p{4.5cm}|p{2.3cm}|} \hline
Layer embedding dimension & 256\\ \hline
Encoder-decoder layers & 3 \\ \hline
Self-attention heads & 8 \\ \hline
Feedforward layer dimension & 512 \\ \hline
\end{tabular}}
\caption{Training parameters and specifications along with Transformer encoders-decoders model parameters}\label{tab:model_training_parameters}
\end{table}
For LSTM training, we adopted the parameters in \cite{bouhsain2020pedestrian_LSTMPaper}, the dimension of the hidden state is 256 and the encoder-decoder layer is 1. Additional parameters for TF-ed architecture, along with training specifications for all datasets are summarized in Table \ref{tab:model_training_parameters}. The Pytorch library was used for implementation.

%\vspace{-6.75mm}
\subsection{Evaluation Metrics}\label{subsec:eval}
%\vspace{-0.12cm}
Average Displacement Error (ADE) and Final Displacement Error (FDE) of bbox center coordinates are used as evaluation metrics for trajectory prediction (Eqns. \eqref{eq:ADE}, \eqref{eq:FDE}), whereas Accuracy (Eqns. \eqref{eq:acc}) is used as a metric for action prediction. ADE measures the general fit of the prediction with regard to the ground truth, averaging the discrepancy at each time step. 
\begin{align}
    ADE &= \frac{1}{T} \sum_{t=1}^T \sqrt{(x_{i}-x_{i}^{GT})^2 + (y_{i}-y_{i}^{GT})^2}, \label{eq:ADE}\\
    FDE &= \sqrt{(x_{T}-x_{T}^{GT})^2 + (y_{T}-y_{T}^{GT})^2},\label{eq:FDE}\\
    Accuracy &= \frac{TP + TN}{TP + TN +FP +FN} \label{eq:acc}
\end{align}
ADE is the root mean square error (RMSE) of all the predictions and ground truths during the forecasting period, while FDE denotes the RMSE between the final prediction at the end of the prediction sequence and the corresponding ground truth. Both ADE and FDE in our case are in pixel units.

\section{Results and Discussion}\label{sec:dis_res}
% \alvika{In this section: check for consistency we use "predicted future time steps","predicted future frames". Its confusing need to stick to one}
\subsection{Detection and Tracking}
 Sample output frames of pedestrian detection and tracking for self-driving shuttle data are shown in Fig. \ref{tracking}, where all pedestrians in the image frame are successfully detected and tracked in Fig. \ref{fig:Tracking success}. Fig. \ref{fig:Tracking failure} shows a failure scenario where the pedestrian in the middle of the road is not detected (inset figures show a closer view of pedestrians).

Intermittent detection failure mentioned above is particularly observed along the path consisting of heavy tree coverage on both sides of the road thus further affecting the already less-than-ideal evening illumination conditions. 
% \alvika{need to revise the next line after completing the introduction}
% Furthermore, the pedestrians ID switch issue from tracking will result in the inconsistency of the observed pedestrians' bounding boxes trajectories, which will affect the down stream action and trajectory prediction performance adversely. 
% Further, the particular stretch of the path shown in Fig. \ref{fig:Tracking failure} or the pedestrian is detected as other classes, hence no detection showing since we only track class person.
%\alvika{I commented out the above since it would have been a false detection even if we were considering other classes, so didnt find this relevant}
\subsection{Action and Trajectory Prediction}
The action prediction accuracy results as well as the ADE and FDE results for trajectory prediction on JAAD and CARLA testing set are summarized in Table \ref{tab:Intent_results_summary1}, where ADE and FDE are calculated in a batch manner. The inference runtime is also reported. We only include LSTM-ed and TF-ed in our evaluation because other models are not comparable directly due to different action definitions, data used, pipelines, etc. Given observations with length $O$, we predict the actions and trajectories for the next $T$ frames, and we refer to this as a \emph{prediction sequence} of length T. Results of $O=16,  T=1$, $O=T=16$ and $O=T=25$ are reported. In particuler, the TF-ed ($T=1$) model uses 1 layer 1 head for training, whereas the others use 3 layers and 8 heads. 

\begin{figure}[!h]
\centering
%\hspace{-0.2in} lbrt
\begin{subfigure}{0.45\textwidth}
    \includegraphics[width=\textwidth,trim=2cm 7cm 3cm 6cm, clip=true]{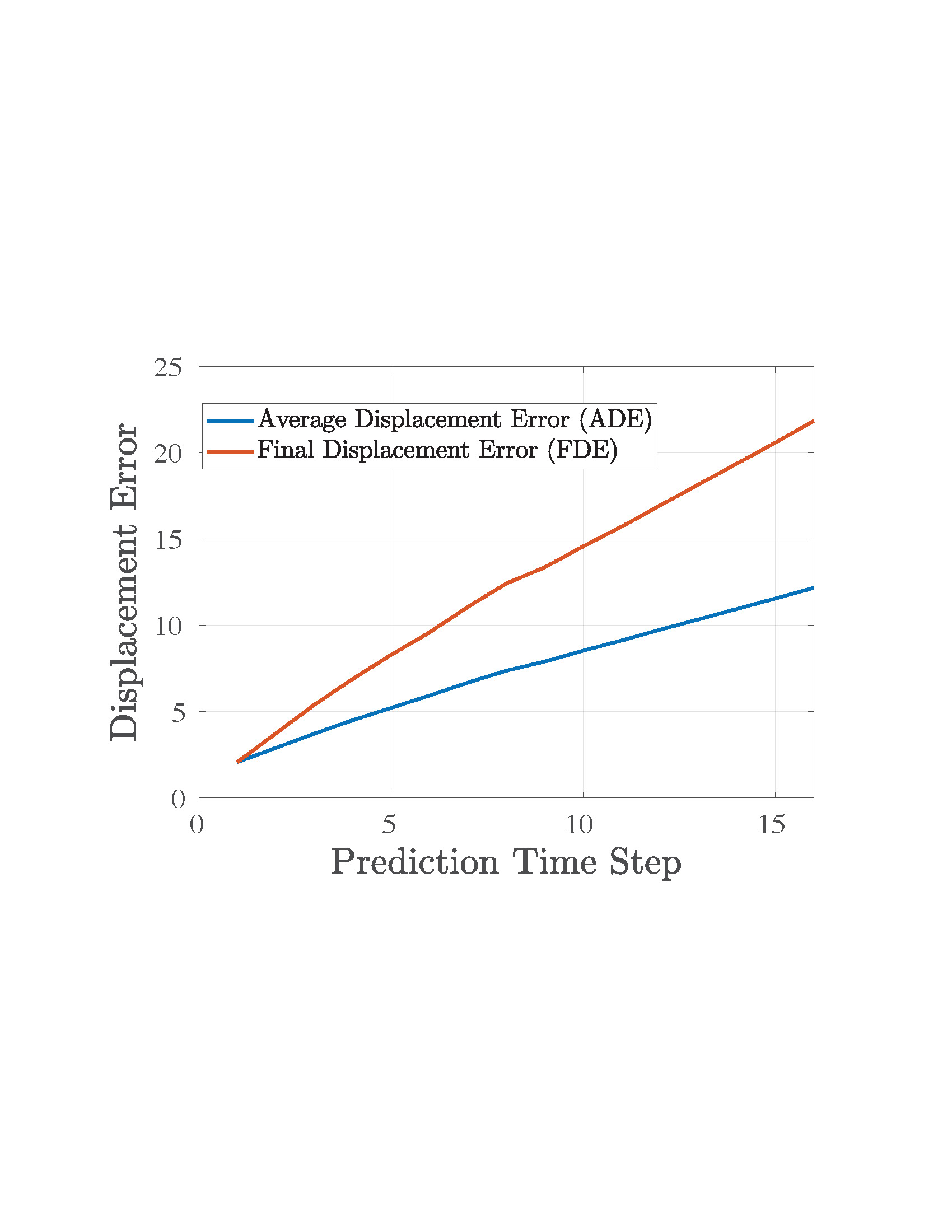}
   
    \caption{ADE and FDE}
     \label{fig:ade_fde}
 \end{subfigure}
\hfill
\begin{subfigure}{0.5\textwidth}
    \includegraphics[width=\textwidth,trim=2cm 7.5cm 3cm 8cm, clip=true]{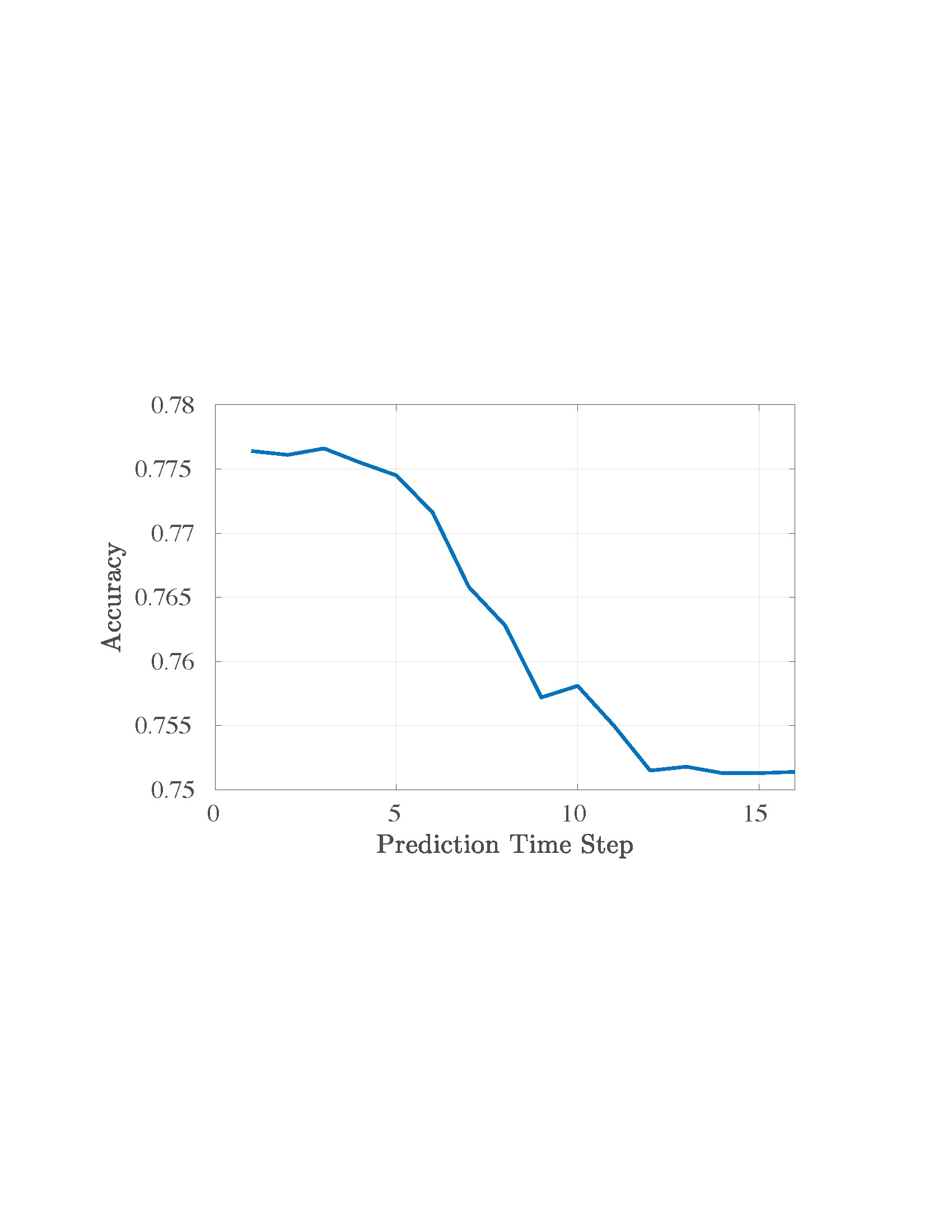}

    \caption{Accuracy}
    \label{fig:accuracy}
\end{subfigure}
\caption{ADE, FDE, and Accuracy change with different predicted time step from 1 to 16 given the same observations from JAAD testing set for TF-ed model }
\label{fig:ade}
\end{figure}

\begin{figure*}[!h]
%\centering

\begin{subfigure}{\textwidth}
\includegraphics[width=0.24\textwidth]{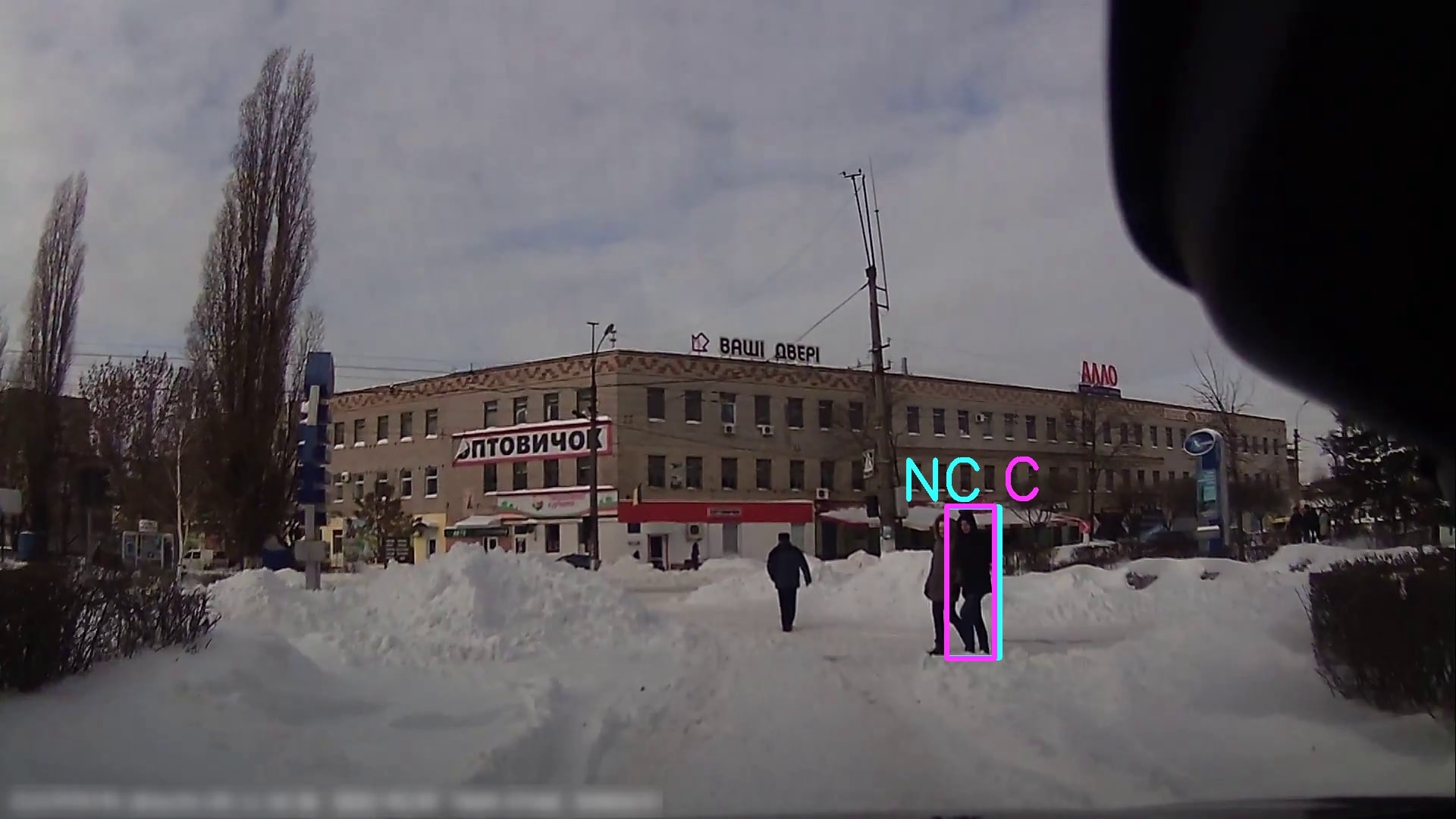} \hfill
\includegraphics[width=0.24\textwidth]{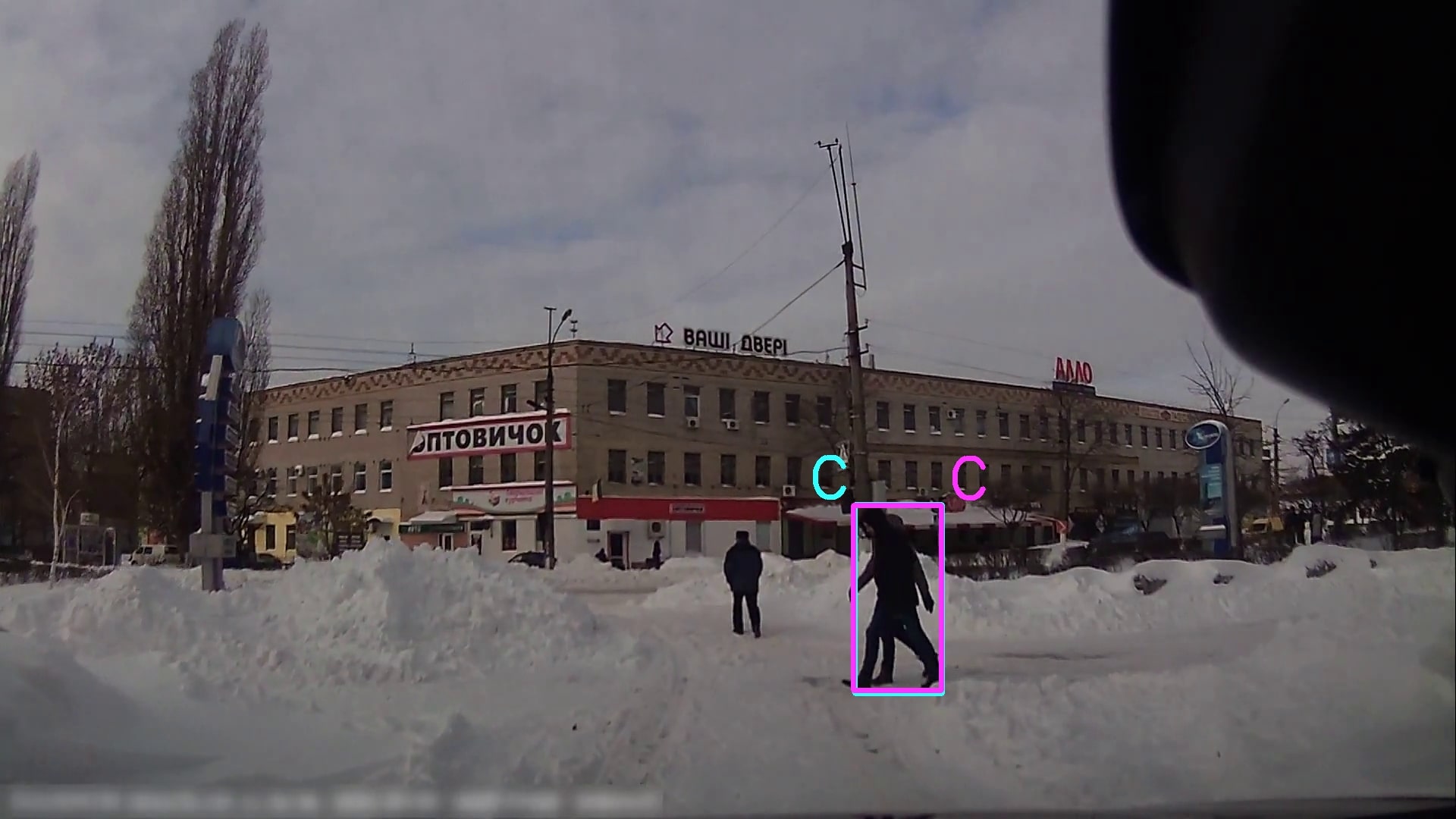} \hfill
\includegraphics[width=0.24\textwidth]{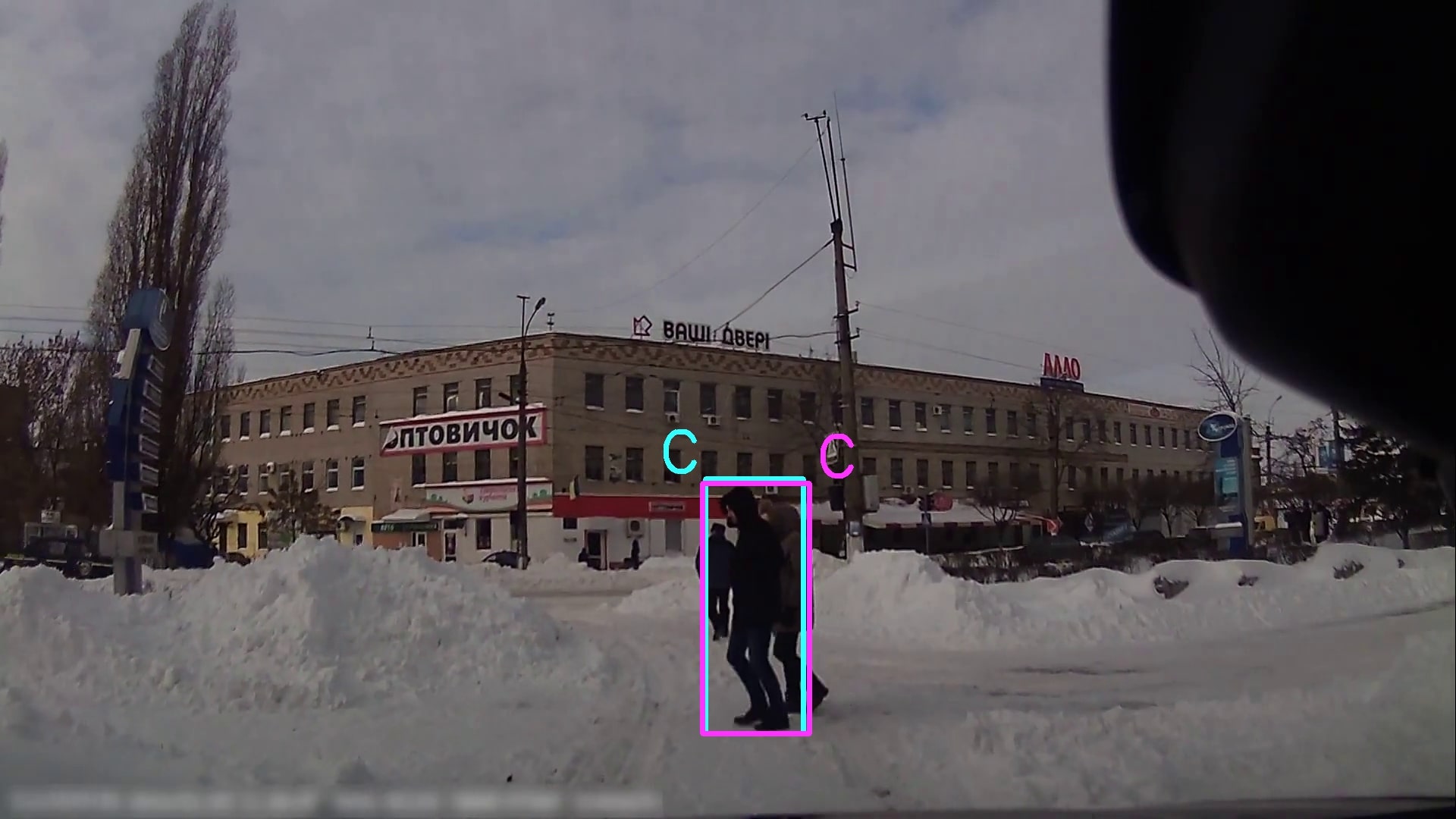} \hfill
\includegraphics[width=0.24\textwidth]{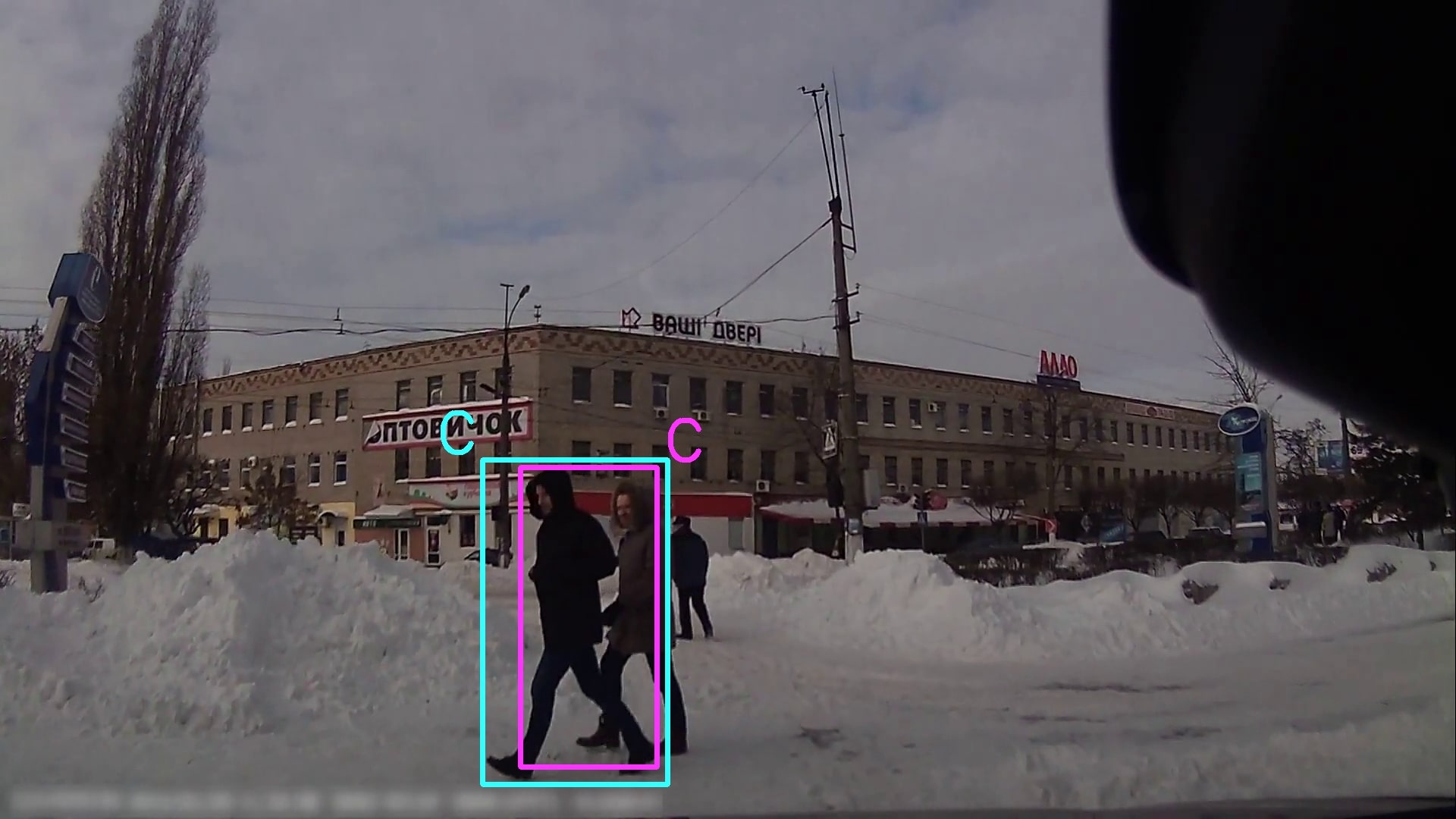} 
\caption{JAAD Results}
\label{fig:JAAD examples}
\end{subfigure}

\begin{subfigure}{\textwidth}
\includegraphics[width=0.24\textwidth]{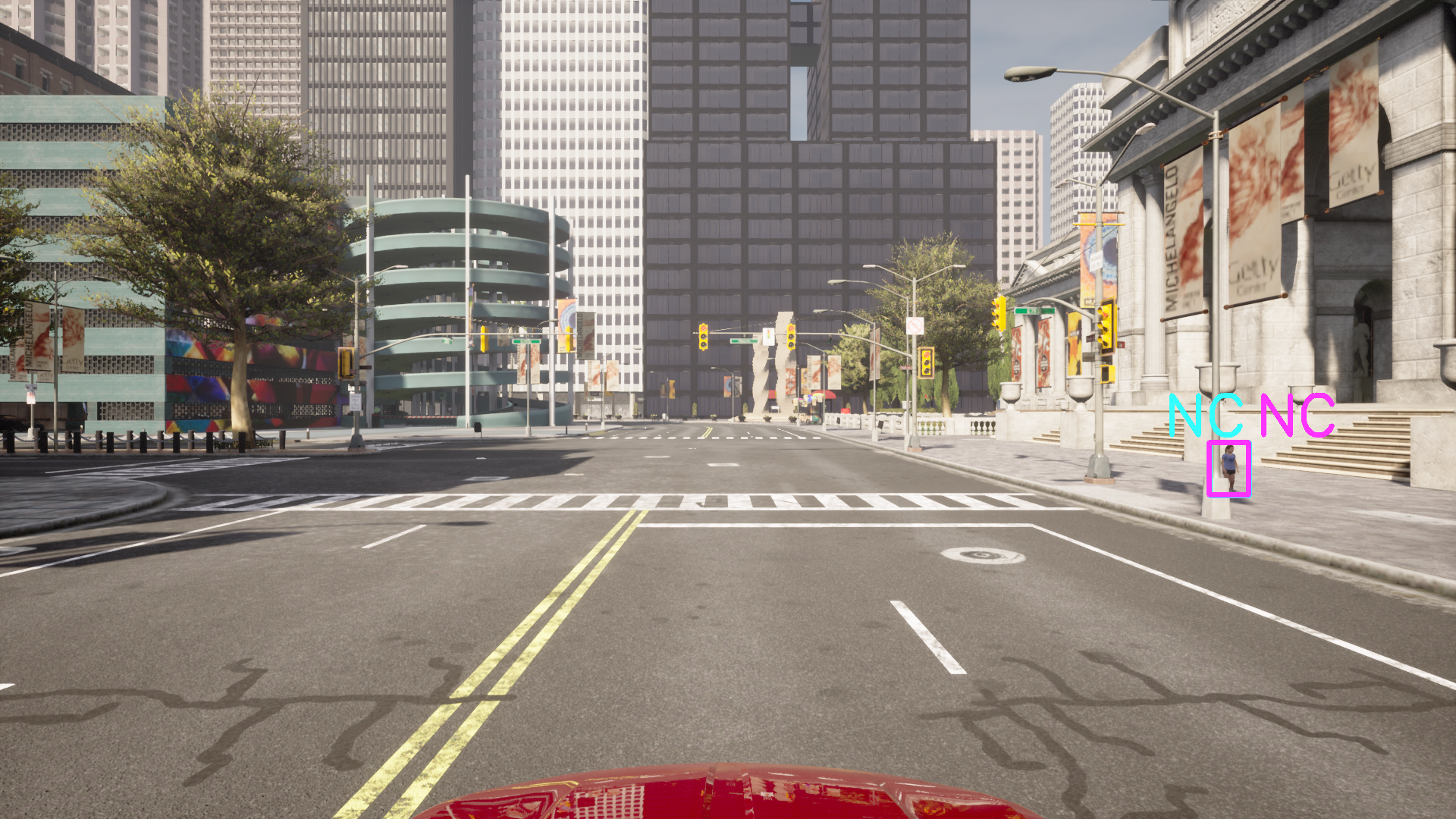} \hfill
\includegraphics[width=0.24\textwidth]{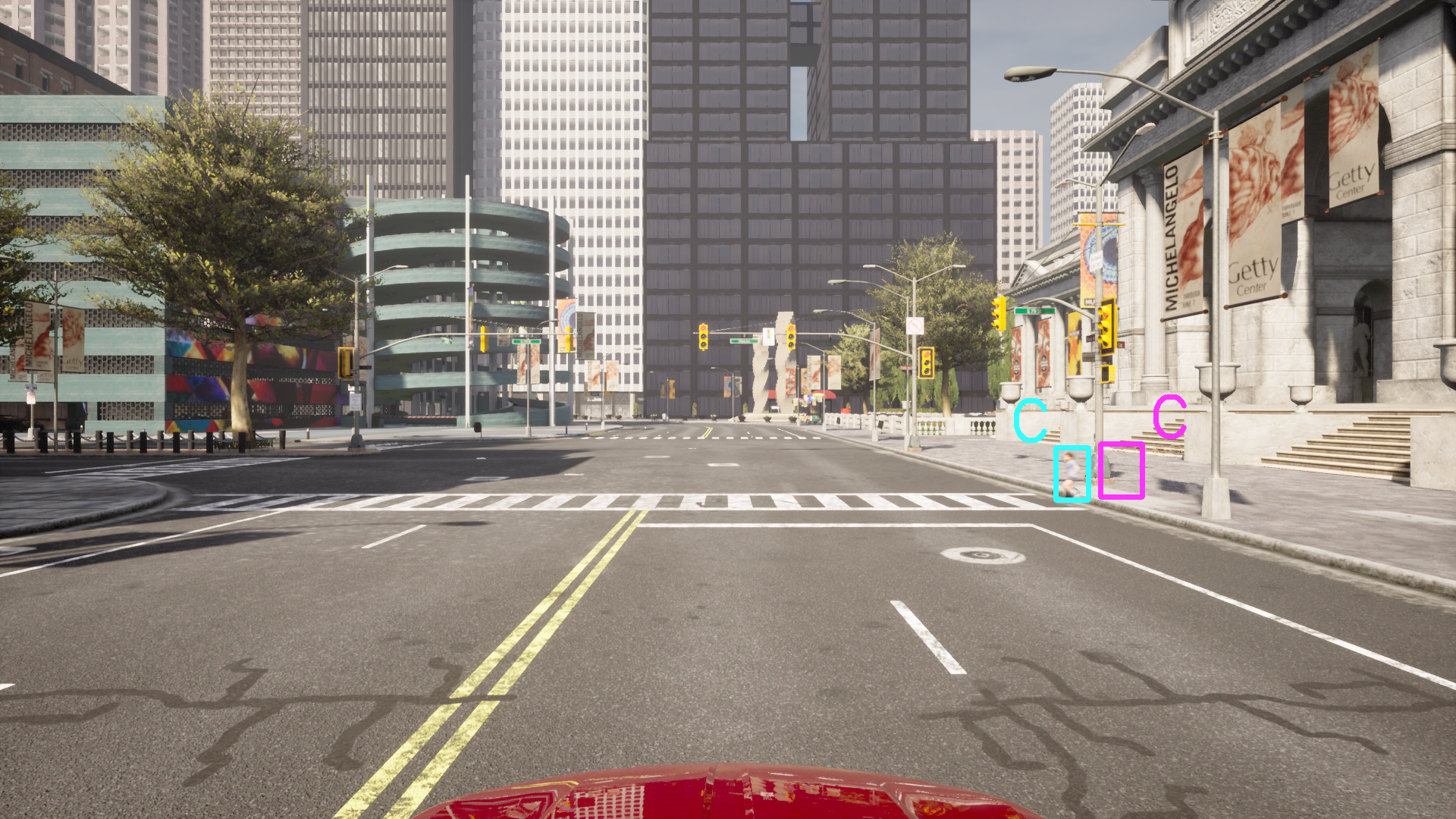} \hfill
\includegraphics[width=0.24\textwidth]{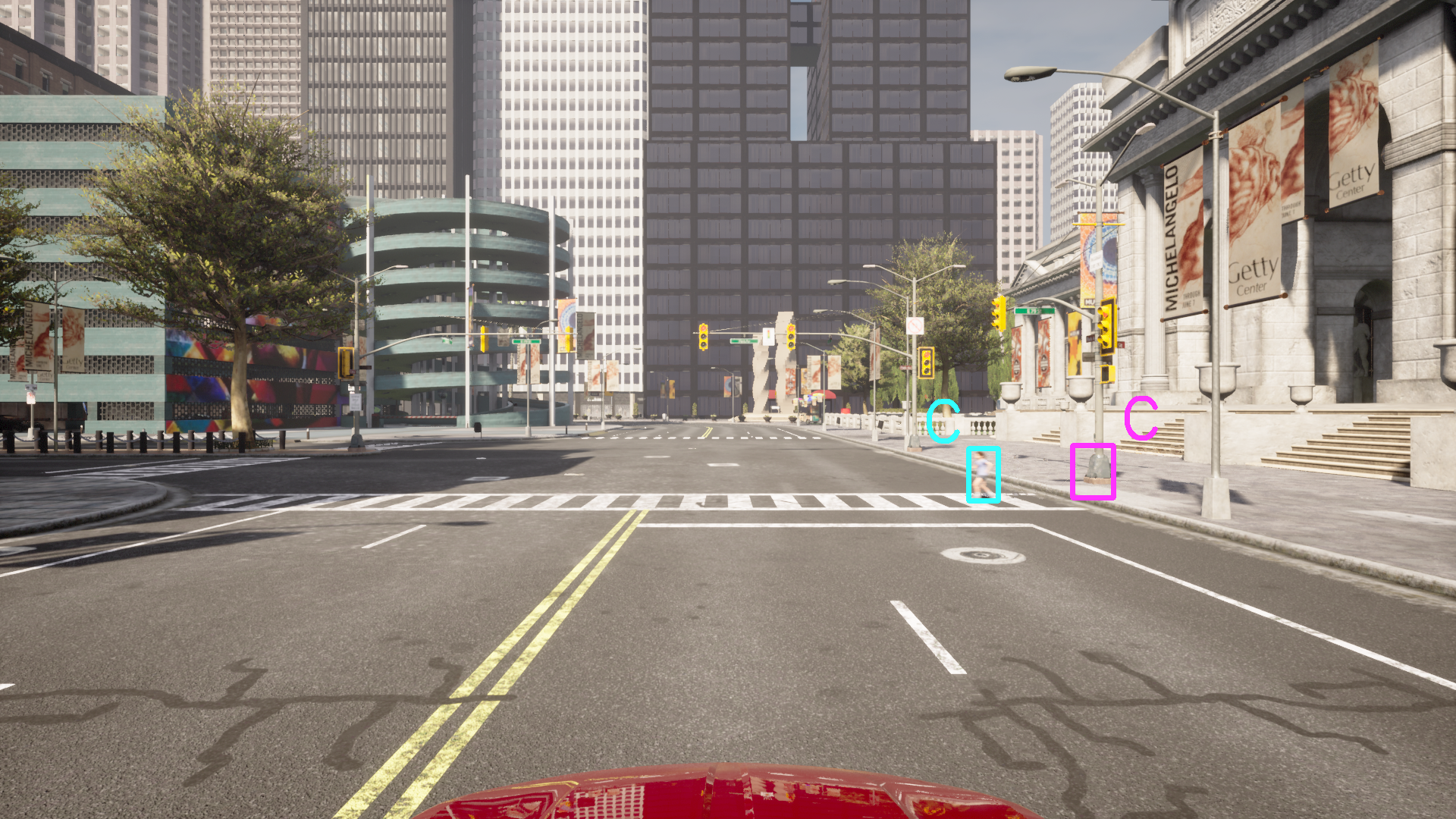} \hfill
\includegraphics[width=0.24\textwidth]{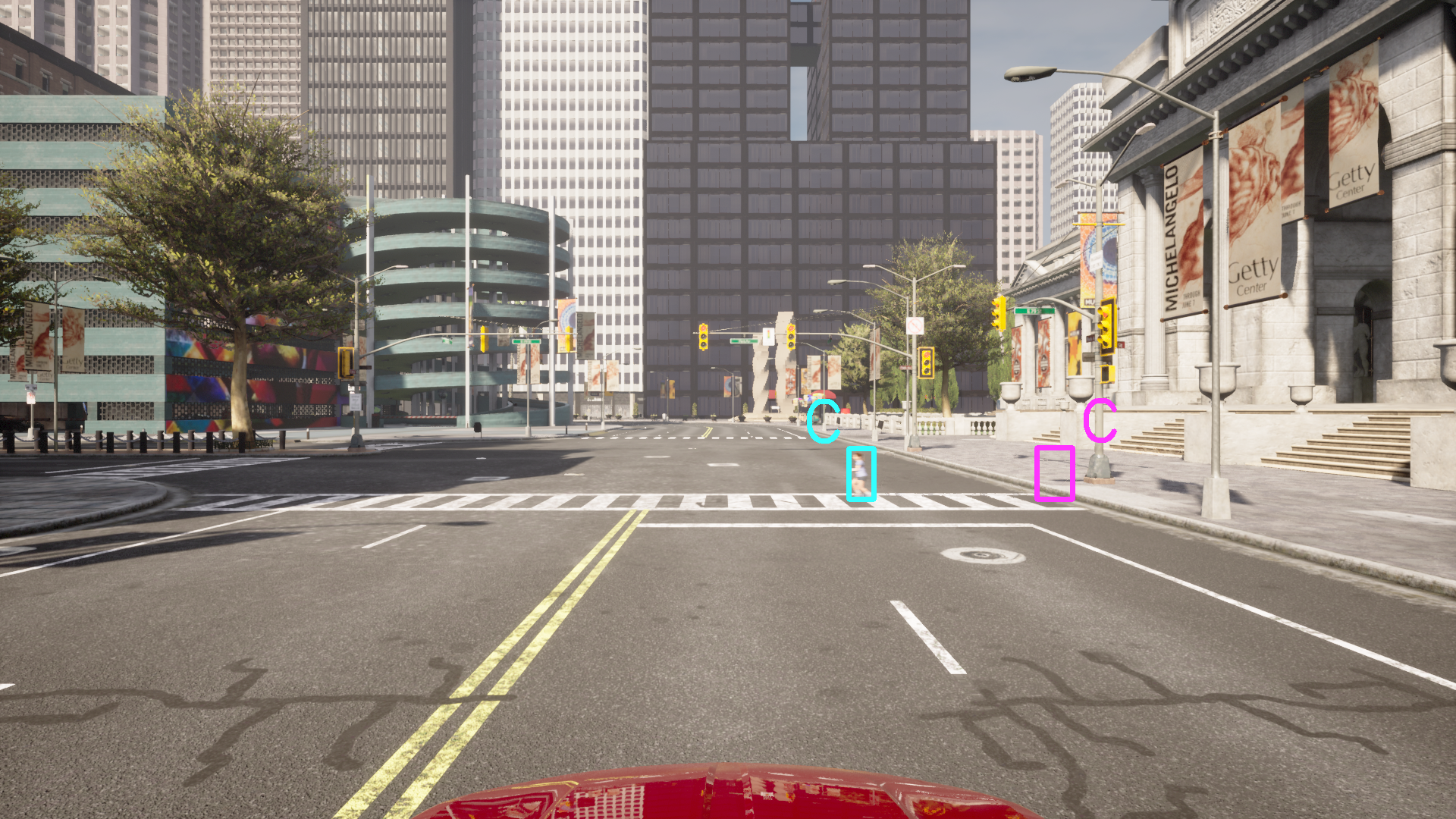} \hfill
\caption{CARLA results}
\label{fig:CARLA examples}
\end{subfigure}

\begin{subfigure}{\textwidth}
\includegraphics[width=0.24\textwidth]{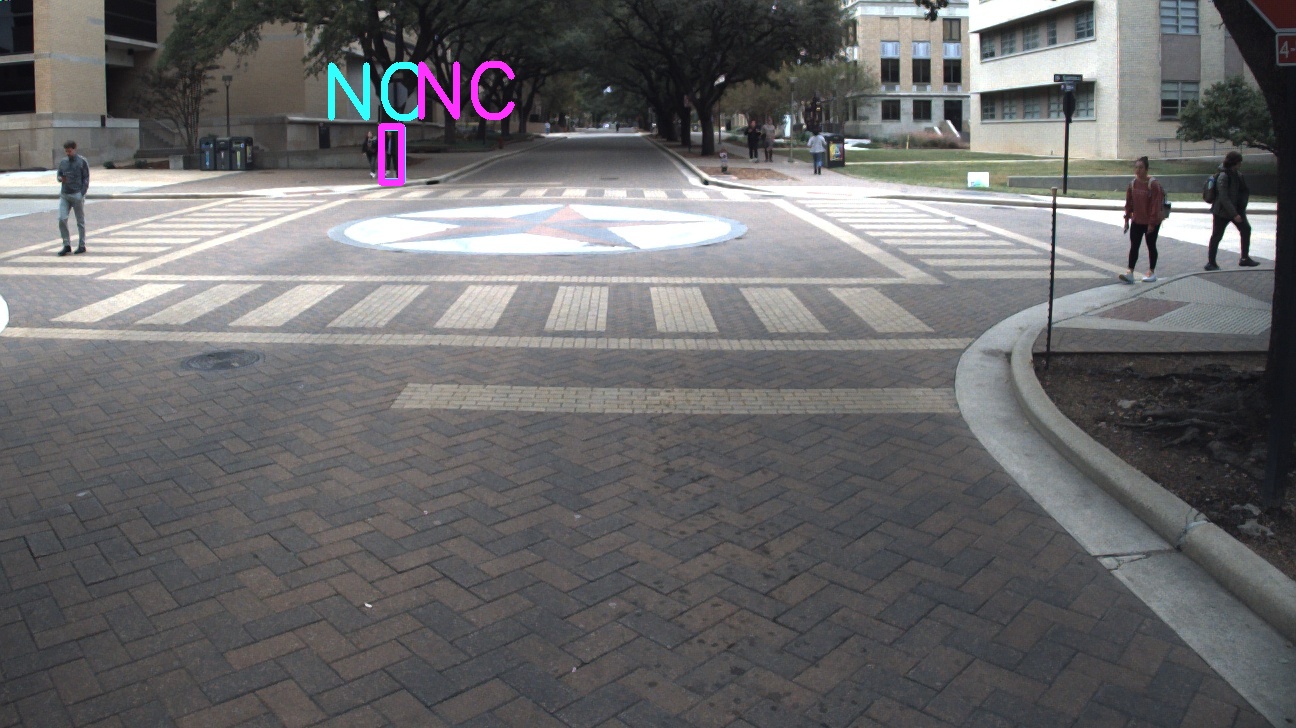} \hfill
\includegraphics[width=0.24\textwidth]{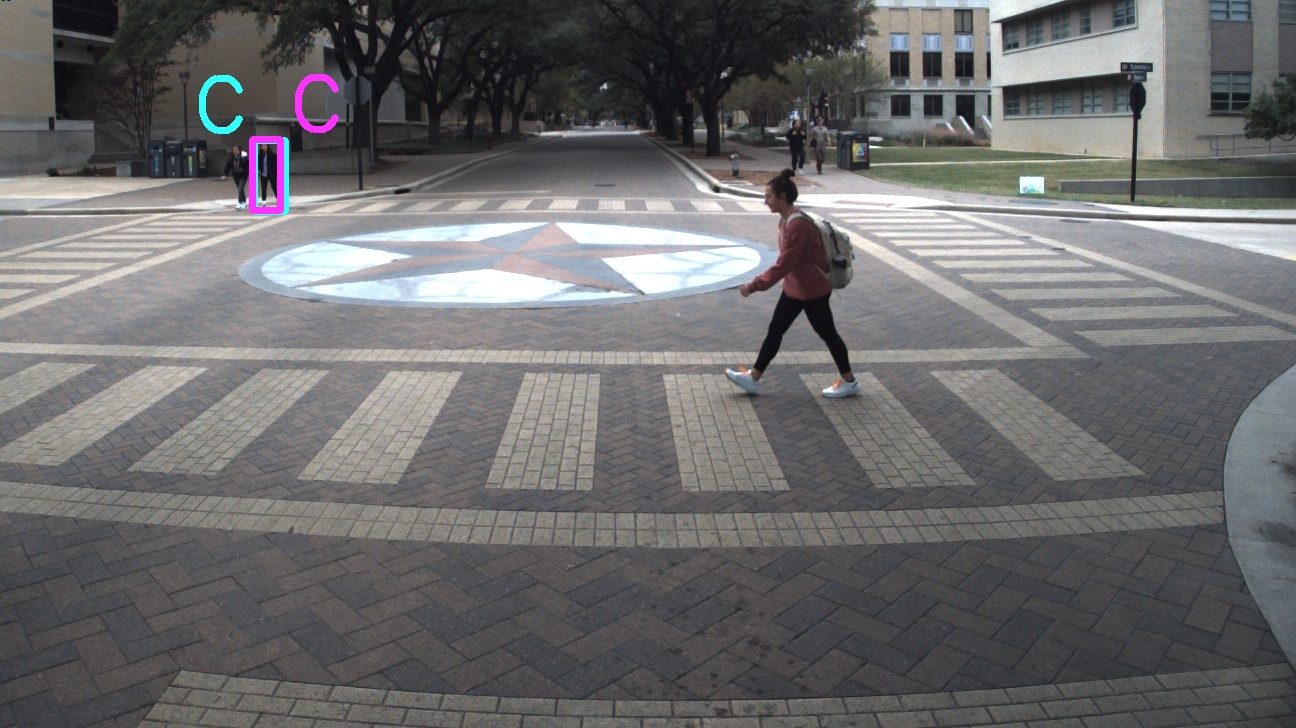} \hfill
\includegraphics[width=0.24\textwidth]{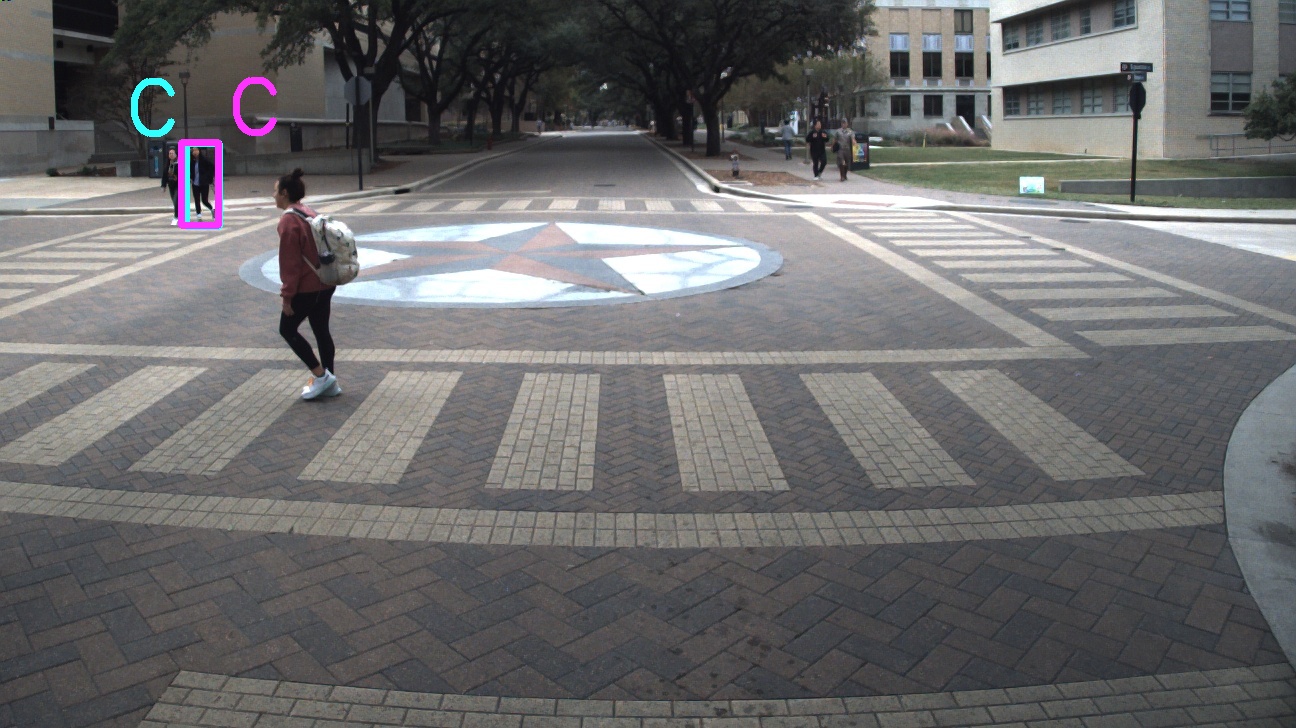} \hfill
\includegraphics[width=0.24\textwidth]{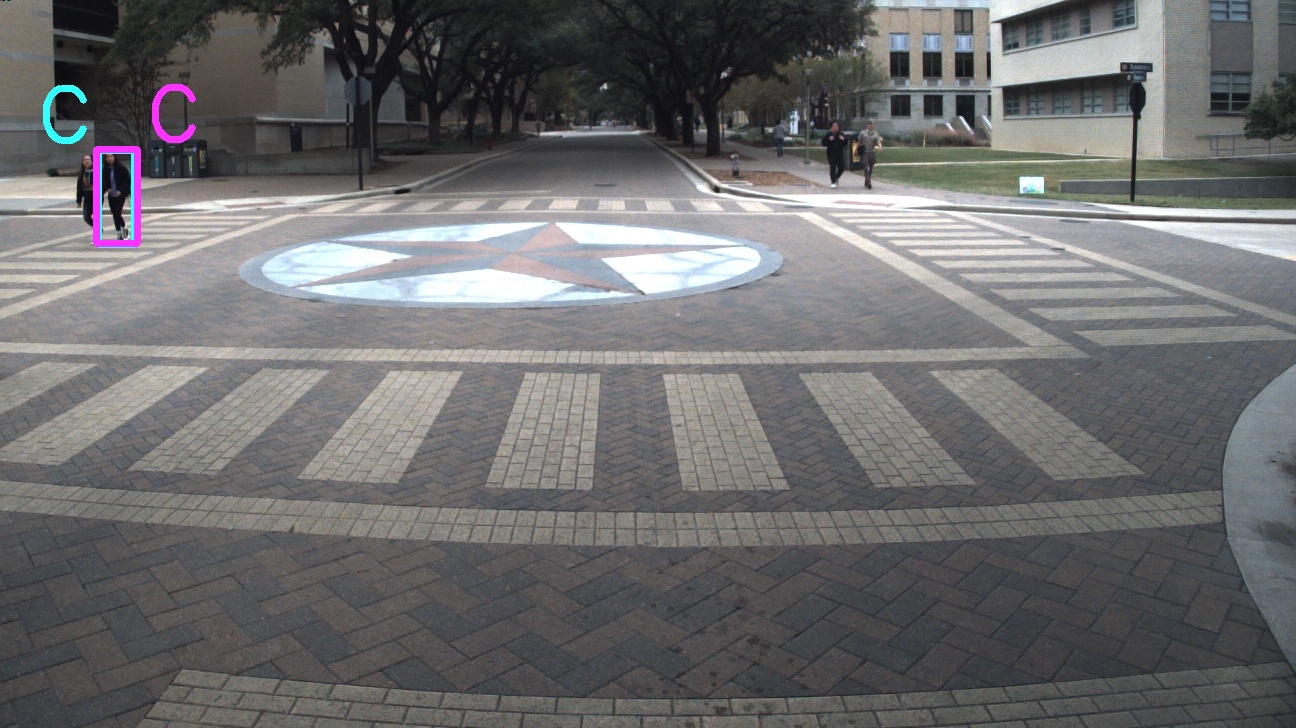} \hfill
\caption{Shuttle data results set 1}
\label{fig:Shuttle examples 1}
\end{subfigure}

\begin{subfigure}{\textwidth}
\includegraphics[width=0.24\textwidth]{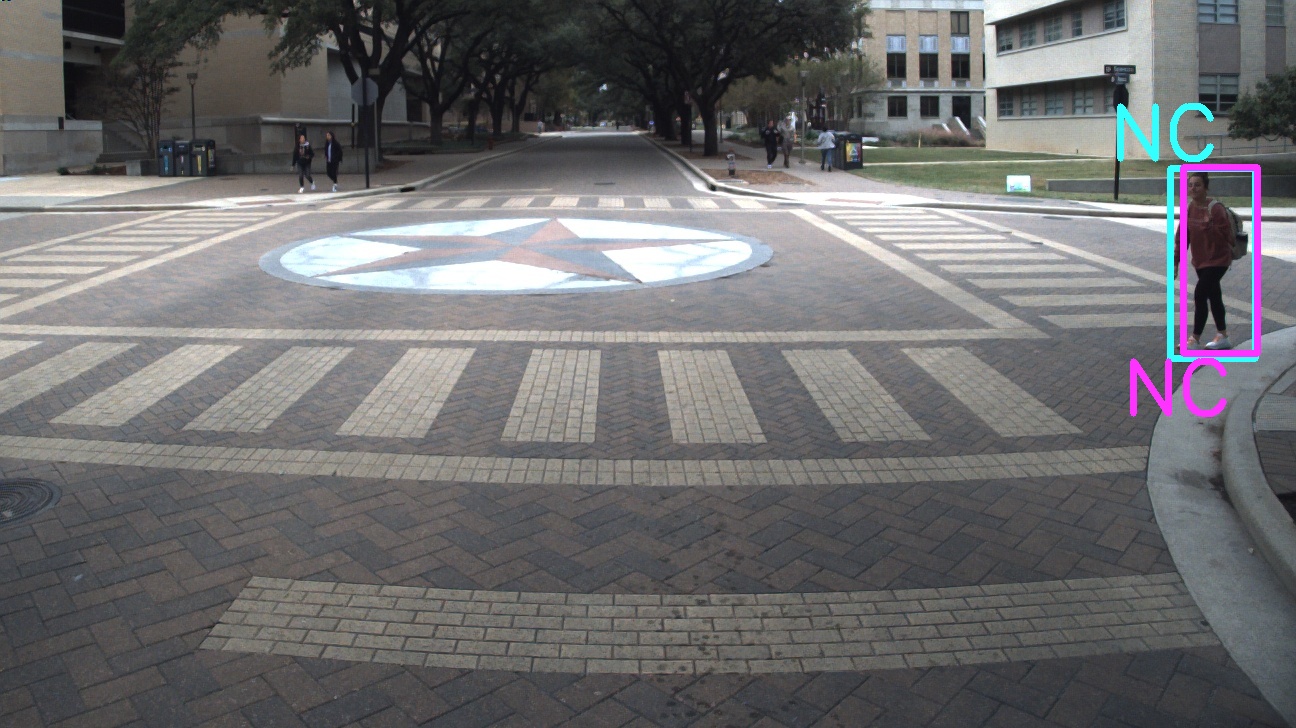} \hfill
\includegraphics[width=0.24\textwidth]{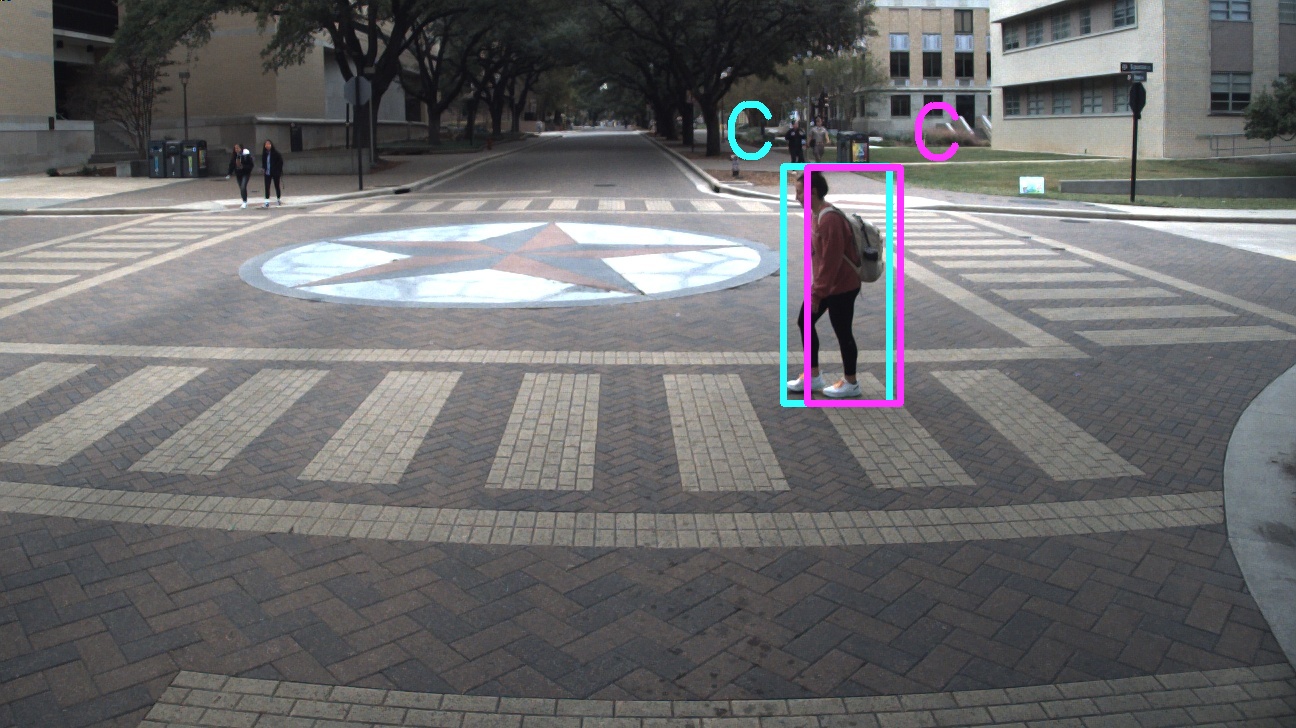} \hfill
\includegraphics[width=0.24\textwidth]{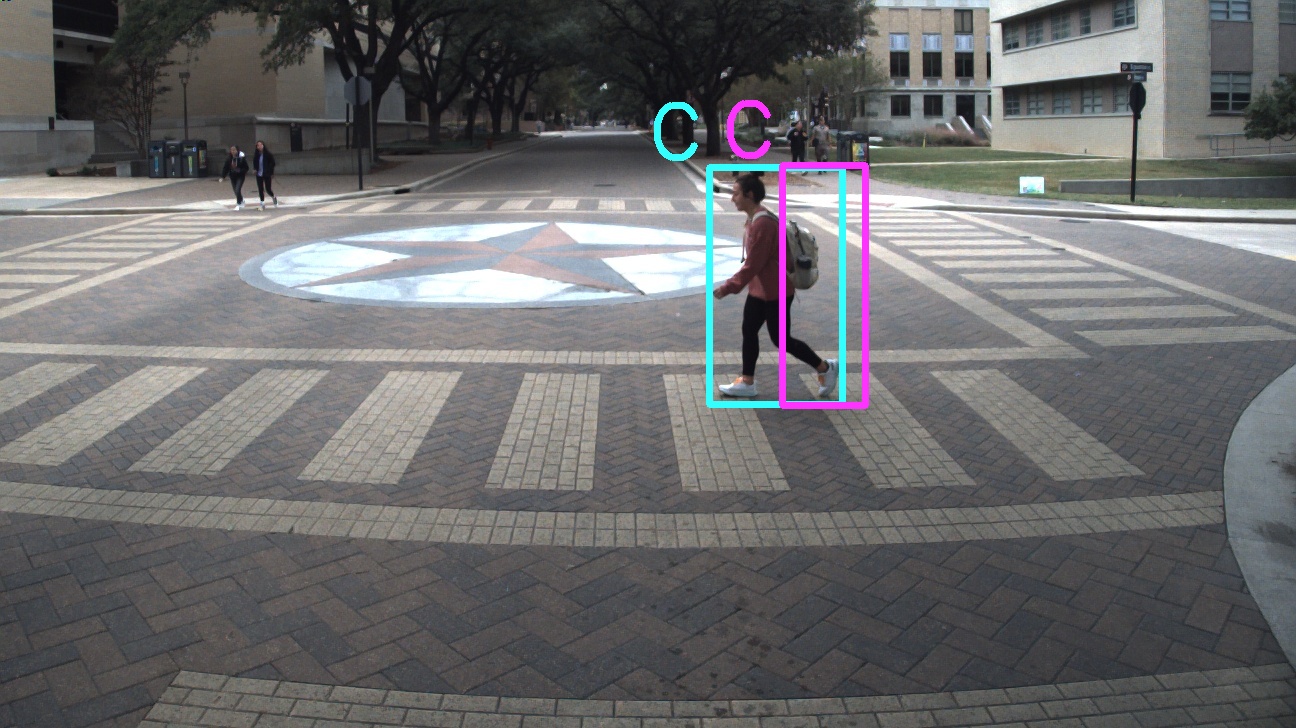} \hfill
\includegraphics[width=0.24\textwidth]{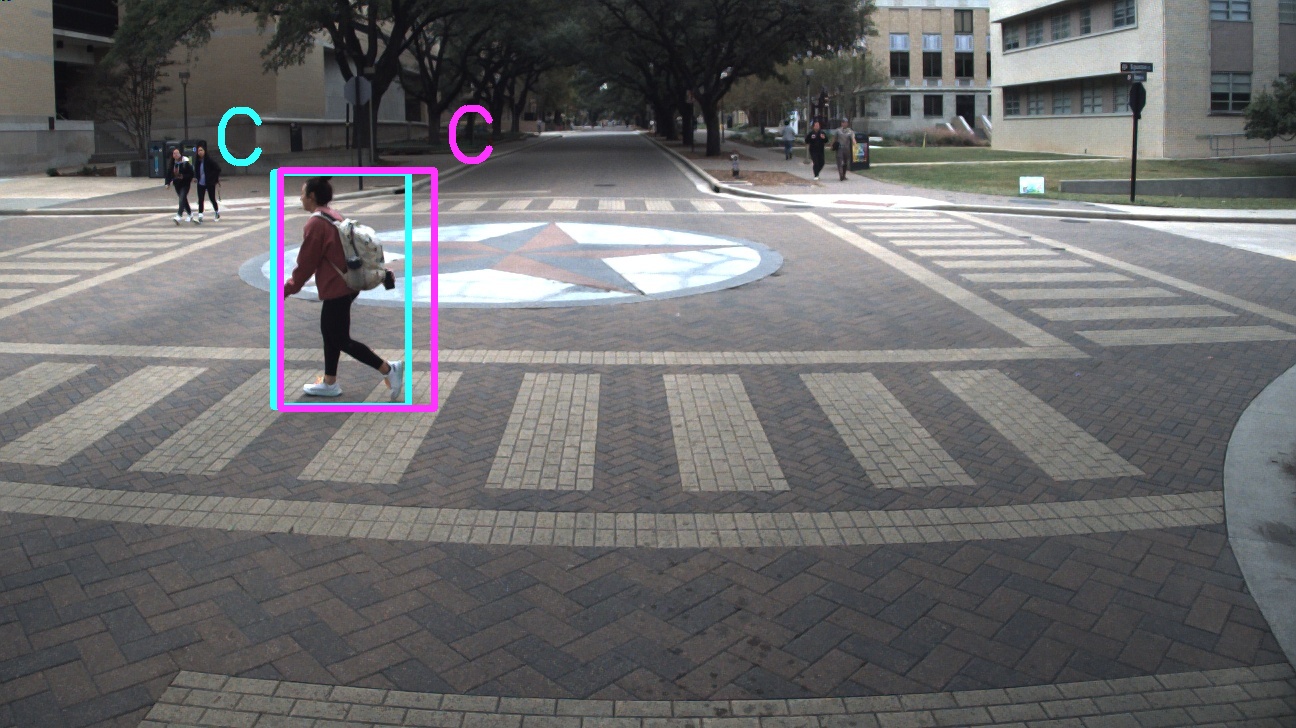} 
\hfill
\caption{Shuttle data results set 2}
\label{fig:Shuttle examples 2}
\end{subfigure}

\begin{subfigure}{\textwidth}
\includegraphics[width=0.24\textwidth]{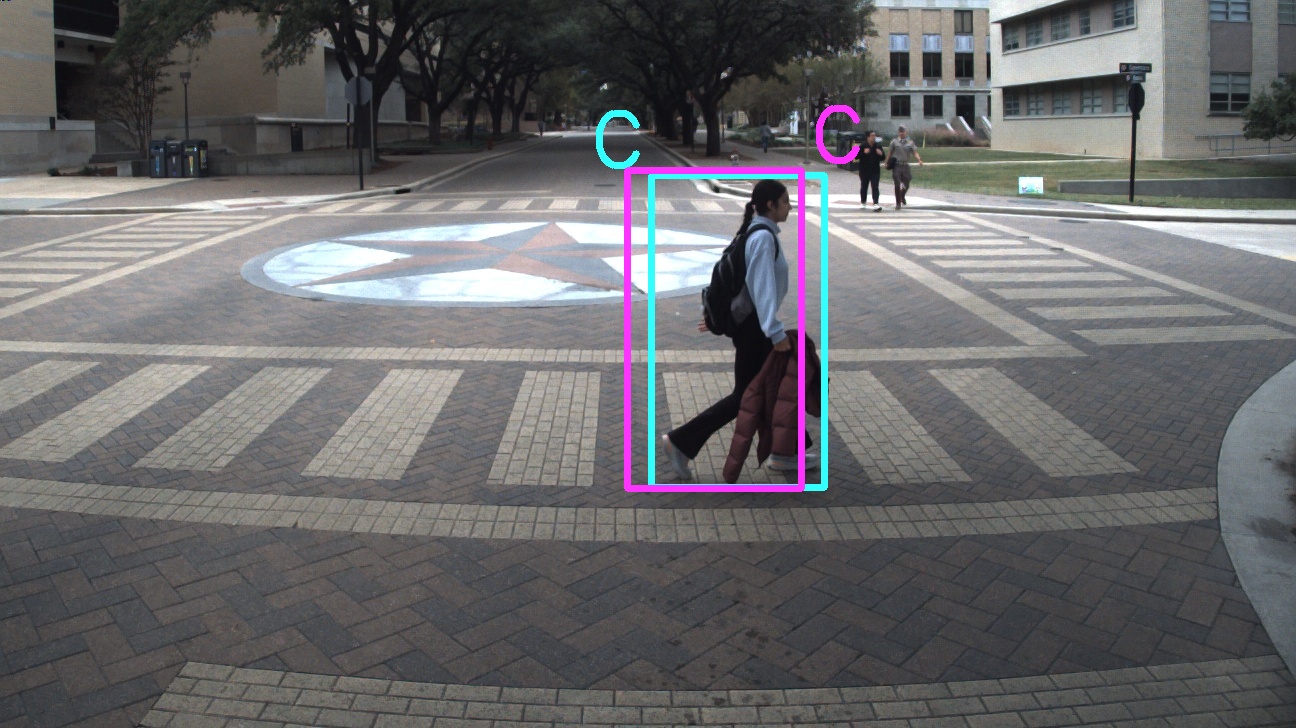} \hfill
\includegraphics[width=0.24\textwidth]{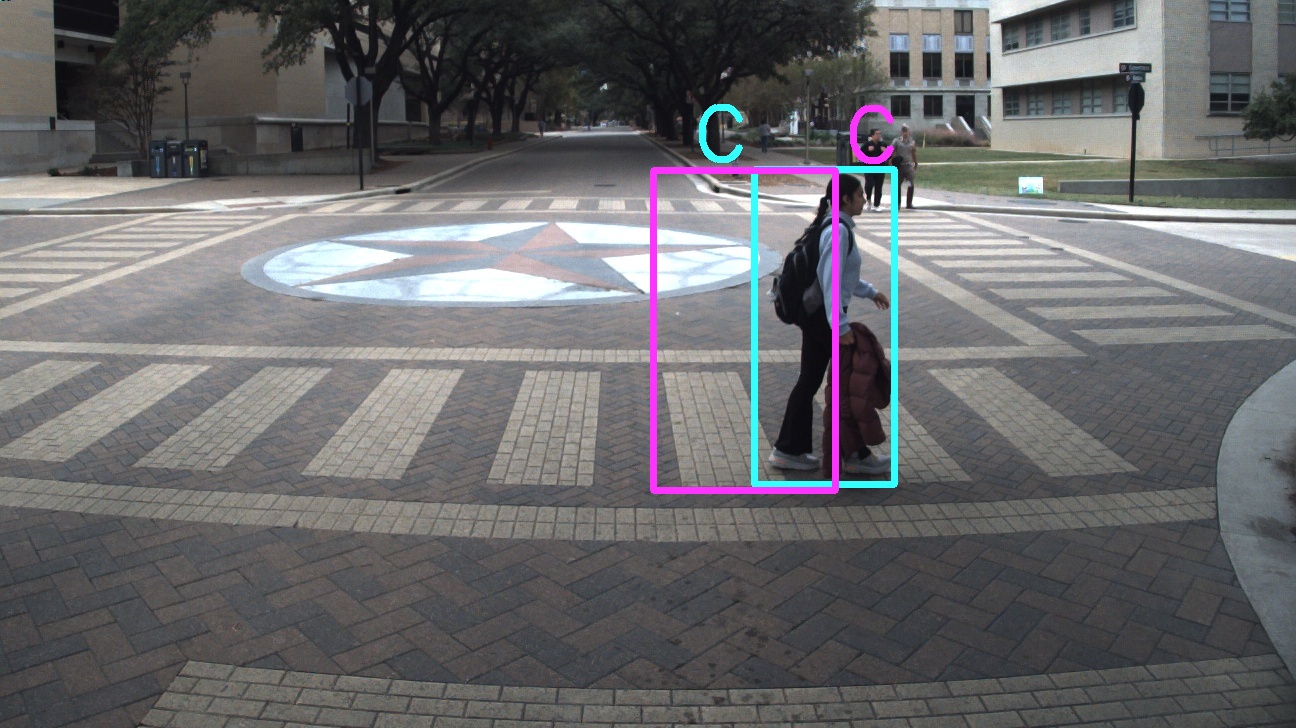} \hfill
\includegraphics[width=0.24\textwidth]{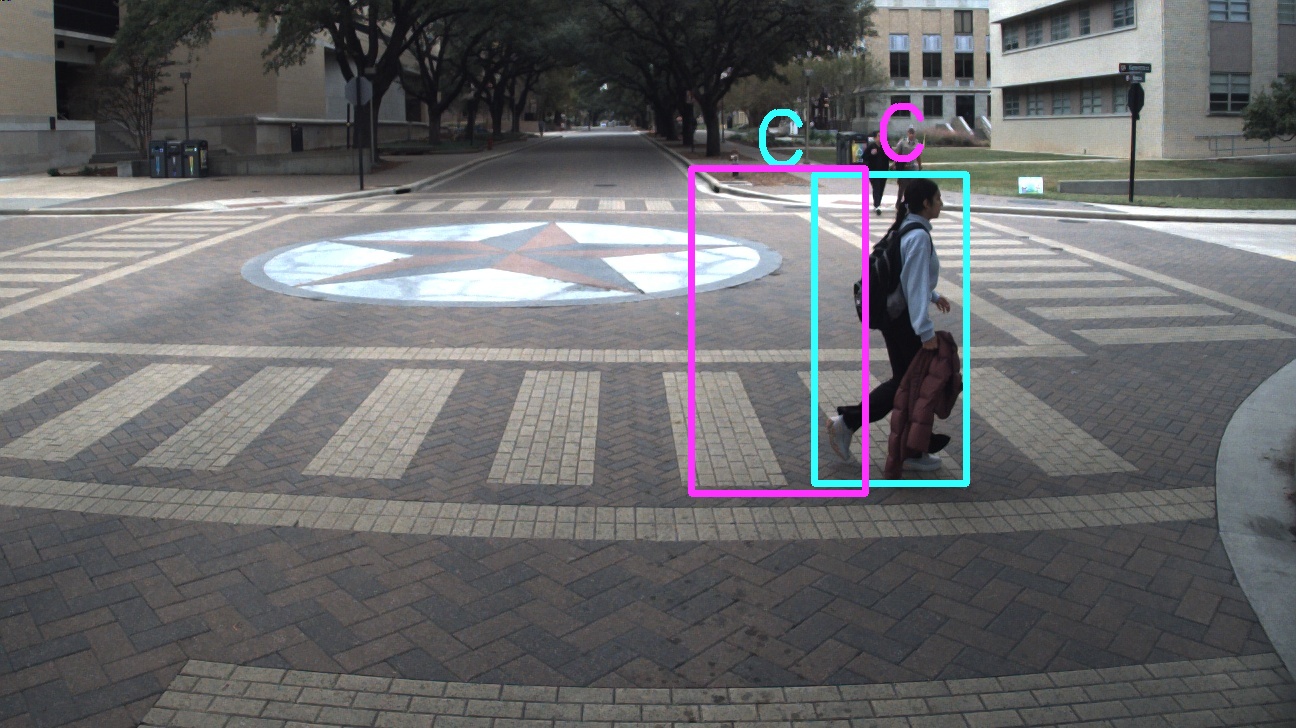} \hfill
\includegraphics[width=0.24\textwidth]{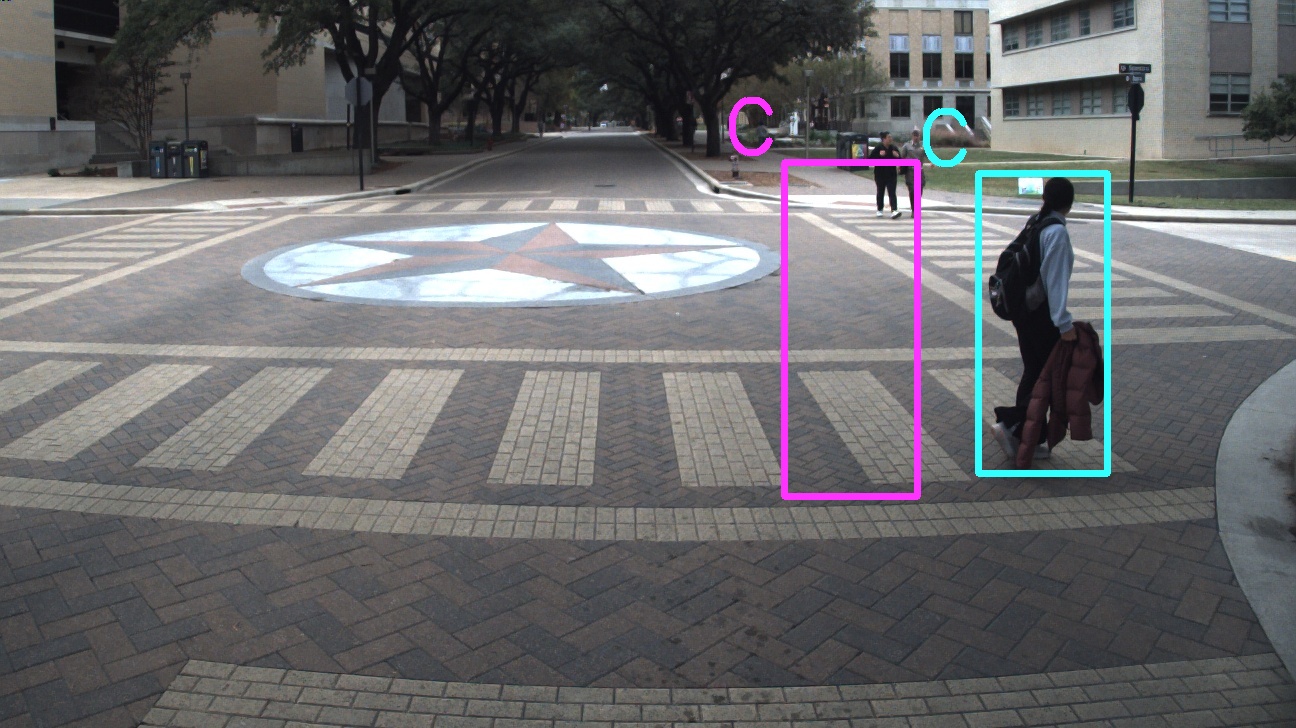} \hfill
\caption{Shuttle data results set 3}
\label{fig:Shuttle examples 3}
\end{subfigure}
\caption{Visualization of the action and trajectory prediction examples obtained by TF-ed on JAAD testing data, CARLA simulation data and self-driving shuttle real date. Ground Truth (GT) and predicted bboxes are in cyan and pink color, respectively. C denotes crossing while NC denotes not crossing}
\label{fig: result examples}
\end{figure*}

For JAAD testing set, the TF-ed model outperforms the LSTM-ed for both tasks when making only one prediction, where ADE and FDE have been decreased by 14.4 \% and accuracy has increased by nearly 4\%. With the prediction sequence length increasing to 16 and 25, LSTM-ed has lower ADE and FDE but less accurate than TF-ed for action prediction. In particular, the action prediction accuracy improves by 7.4 \% when $T=25$ by TF-ed.  For CARLA data evaluation, LSTM-ed has much better performance  in all cases.  In general, the runtime of LSTM-ed is much less than that of TF-ed. Fig. \ref{fig: result examples} shows 5 sets of action and trajectory prediction examples from three evaluation scenarios with time steps from left to right for TF-ed model. In Fig. \ref{fig:JAAD examples}, a pedestrian is waiting at the side of the road first, which lasts for a few seconds, and starts to cross in the second frame and the vehicle is approaching at the same time. Fig. \ref{fig:CARLA examples} simulates a pedestrian crossing at a farther distance while the ego vehicle stops behind the crosswalk at a distance. As seen in the results, although the action predictions are correct, ADE and FDE are gradually increasing proportionally with time step of the prediction sequence. This is also observed in self-driving shuttle data results shown in Fig. \ref{fig:Shuttle examples 3}. This can be explained by the trend shown in Fig. \ref{fig:ade} which shows an increasing accumulated error with increase in prediction time steps. In other words, the first and the last predicted frame of each predicted sequence has the lowest and the highest ADE/FDE respectively. It should be noted that the error offset is most significant in Figs. \ref{fig:Shuttle examples 3} as compared to Figs. \ref{fig:JAAD examples}, and \ref{fig:Shuttle examples 2}. This is because only Fig. \ref{fig:Shuttle examples 3} includes frames sampled from the same prediction sequence thus reflecting a more significant effect of gradually increasing error, whereas the image frames for other examples are the first prediction samples from four different prediction sequences to include a longer pedestrian trajectory and thus do not show a gradually increasing error offset. Further, we observe that for Fig. \ref{fig:Shuttle examples 3} pedestrian is not detected and tracked robustly in the first few seconds of the motion which explains the delay in the prediction of actions.

\begin{comment}
   \hfill
\begin{subfigure}{0.2\textwidth}
    \includegraphics[width=\textwidth]{00041.png}
\end{subfigure}
\hfill
\begin{subfigure}{0.2\textwidth}
    \includegraphics[width=\textwidth]{00043.png}
\end{subfigure}
\hfill
\begin{subfigure}{0.2\textwidth}
    \includegraphics[width=\textwidth]{00045.png}
\end{subfigure} 
\end{comment}

\begin{table}[!ht]
\noindent
\begin{subtable}{1\textwidth}
\centering
\resizebox{\columnwidth}{!}{
\begin{tabular}{ccccc} \hline
{\bf } & \multicolumn{4}{c}{\bf JAAD}\\\hline
        % HMM & TBA& TBA & TBA \\
        {\bf } & ADE(pixel) &FDE(pixel) & Accuracy & Runtime(ms) \\
         LSTM-ed (T=1) &2.08 &2.08 & 0.78 & 5.5 \\
         LSTM-ed (T=16) &12.17 & 21.83 & 0.75 &13.7 \\
         LSTM-ed (T=25) &20.26& 40.57&   0.75 & 17.2  \\
         TF-ed (T=1) & \textcolor{blue}{1.78} &\textcolor{blue}{1.78}& \textcolor{red}{0.81} & 36.8  \\
         TF-ed (T=16) & 15.70& 30.26& 0.80& 447.9  \\
         TF-ed (T=25) & 29.36  & 60.55&\textcolor{red}{0.81}  & 838.2  \\
         
          \hline
\end{tabular}}
\caption{JAAD testing results}\label{tab:JAAD results}
\end{subtable}

\begin{subtable}{1\textwidth}
\centering
\resizebox{\columnwidth}{!}{
\begin{tabular}{ccccc} \hline
{\bf } & \multicolumn{4}{c}{\bf CARLA}\\\hline
        % HMM & TBA& TBA & TBA \\
        {\bf } & ADE(pixel) &FDE(pixel) & Accuracy & Runtime(ms)\\
        LSTM-ed (T=1) & \textcolor{blue}{4.01} &\textcolor{blue}{4.01} & \textcolor{red}{0.88}  & 3.6 \\
        LSTM-ed (T=16) &61.82& 123.90&0.82& 12.2\\
        LSTM-ed (T=25) & 161.92& 291.11& 0.65  & 16.9 \\
        TF-ed (T=1) &5.51&5.51 & 0.65&17.9\\
         TF-ed (T=16) & 77.28&148.28 & 0.62  &215.3   \\
         TF-ed (T=25) & 183.07  &352.51 &  0.54 & 344.9 \\ 
          \hline
\end{tabular}}
\caption{CARLA testing results}\label{tab:CARLAresults}
\end{subtable}
       \caption{Action and trajectory prediction comparison for TF-ed and LSTM-ed architectures. T denotes number of frames in a prediction sequence. Highest accuracy is highlighted in red while lowest ADE/FDE are in blue throughout the table.}  \label{tab:Intent_results_summary1}
\end{table}

\begin{comment}
\begin{table}[ht]
\noindent
\small
\resizebox{\columnwidth}{!}{
\begin{tabular}{ccccc} \hline
{\bf Observation-Length}  & {\bf 16 } & {\bf 25} & {\bf 32} \\ \hline
         LSTM &  0.7709 & 0.7502 & 0.7668 \\
         Transformer & 0.7975 & 0.8058 & 0.7401 \\
          \hline
    \end{tabular}}
       \caption{ction prediction accuracy comparison for LSTM
and TF model architecture.T denotes number of prediction
frames}  \label{tab:Intent_results_summary2}
\end{table}  
\end{comment}

\begin{figure}
\centering
\begin{subfigure}{0.45\textwidth}
    \includegraphics[width=\textwidth]{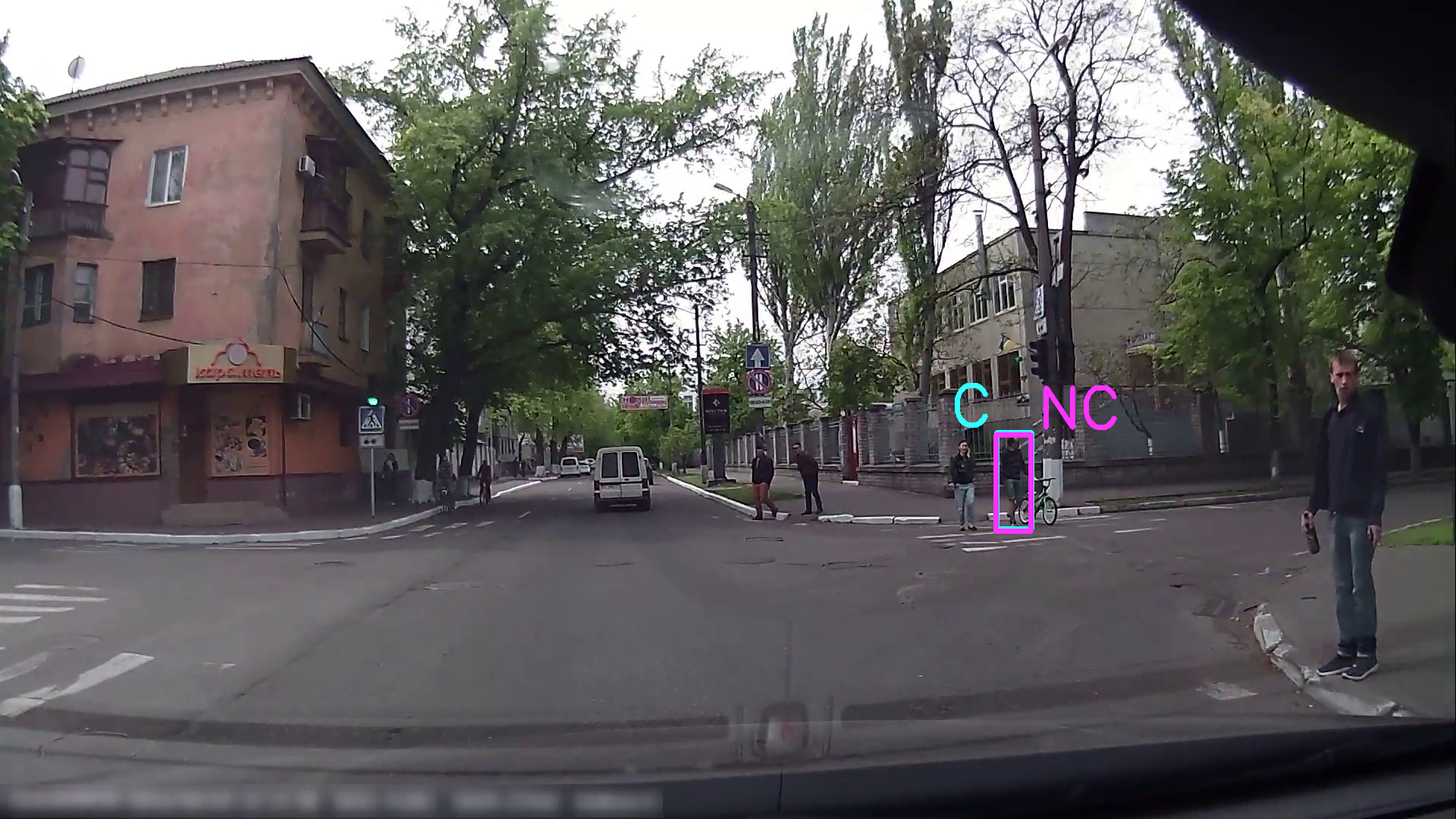}
        \caption{Case a): Failure scenario, JAAD testing data}
        \label{fig:parallel cross}
\end{subfigure}
\hfill
\begin{subfigure}{0.45\textwidth}
    \includegraphics[width=\textwidth]{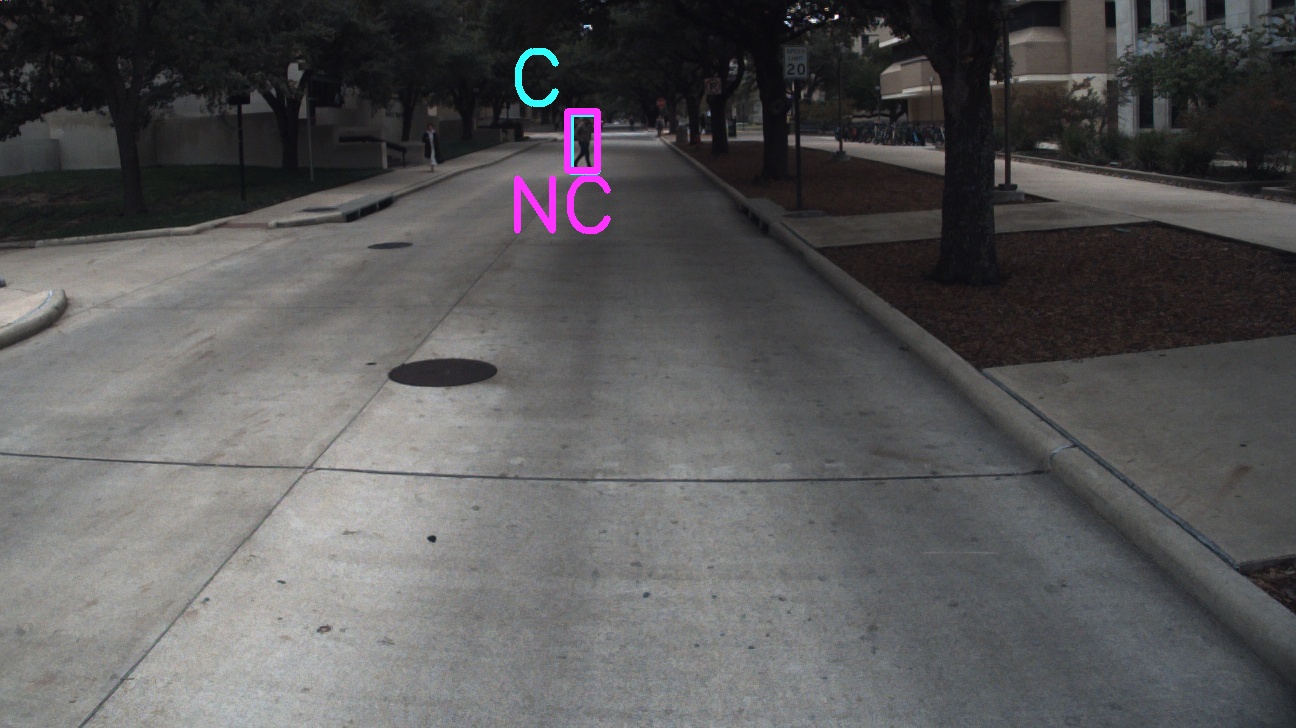}
        \caption{Case b): Failure scenario, Self-driving shuttle data}
        \label{fig:middle} 
\end{subfigure}
\caption{Failure cases of action prediction on JAAD and self-driving shuttle testing data.}
\label{fig: Failure cases}
\end{figure}

To summarize, with only bboxes coordinates as input, LSTM-ed model performs better on pedestrian trajectory prediction task, while the TF-ed model has better performance on pedestrian action prediction task when the prediction sequence length increases. 
Further, we also observed some interesting scenarios which are worth discussing to investigate the efficacy and limitations of pedestrian action prediction methods for safe autonomous driving. For example,
\begin{comment}
   in Fig.\ref{fig:middle}, the man is standing in the middle of the road, even though he is not crossing at the moment, he is very likely to cross in the next second. Nevertheless, the GT annotations for the whole video is "NC", which might cause misleading for future action prediction.  
\end{comment}
in Fig. \ref{fig:parallel cross}, there are two pedestrians crossing on the right side of the ego-vehicle rather than in the path of the vehicle. This scenario is frequently encountered by vehicles turning right at an intersection. In Fig. \ref{fig:middle}, a pedestrian is crossing diagonally (without a crosswalk) while the vehicle is approaching. In both cases, the algorithm fails to predict the correct action. 
Further, in real-time end-to-end evaluation, the tracking ID switch is a major challenge that adversely affects the downstream pedestrian action and trajectory prediction tasks. With a limited camera field of view, there is limited buffer time for the vehicle to react timely  before the pedestrian enters the frame and starts to cross because the accumulation of observations takes time. 

These challenges and scenarios emphasize the need for more exhaustive testing and data collection of vehicle-pedestrian interactions, especially in areas like a university campus, and residential areas with non-signalized intersections.  For safe autonomous driving, it is critical that pedestrian behavior prediction methods are evaluated end to end in real-time (e.g.: effect of tracking failures and accuracy on action prediction accuracy), so that an autonomous shuttle or vehicle can predict the short-term actions robustly, even for non-trivial cases where the pedestrian may not directly be in front of the vehicle, may have a slightly different trajectory profile than those observed typically on crosswalks.  
\begin{comment}
  \hfill
\begin{subfigure}{0.32\textwidth}
    \includegraphics[width=\textwidth]{v319-00052.png}
        \caption{GT: NC. Pred: C}
        \label{fig:crosswalk stop}
\end{subfigure}  
\end{comment}
\section{Conclusion}
We present a Transformer encoders-decoders multitask model for pedestrian trajectory-action prediction in urban driving scenarios for safe autonomous driving. The approach is compared against LSTM encoders-decoders model and evaluated on simulation, publicly available JAAD dataset using ground truth data. This is complemented by an end to end pipeline evaluation on a self-driving shuttle data where the framework is integrated with a vision-based pedestrian detection and tracking algorithm. We also provide a taxonomy for pedestrian behaviour characterization and discuss the hierarchical causal relationships between different behaviour levels. 
Through experimental results, we conclude that short-term action prediction methods relying purely on ego vehicle observations have limitations when it comes to autonomous shuttle last mile connectivity driving scenarios. Tracking ID switching when tracking and predicting actions of multiple pedestrians is also a challenge, which leaves the vehicle with very little time for safe decision-making. Towards this, an interesting direction of future work is to explore group action versus individual pedestrian actions and investigate trade-offs with regard to group action prediction versus identifying and predicting actions only for critical pedestrians.
\bibliographystyle{IEEEtran}
\bibliography{References}

\begin{thebibliography}{10}
\providecommand{\url}[1]{#1}
\csname url@samestyle\endcsname
\providecommand{\newblock}{\relax}
\providecommand{\bibinfo}[2]{#2}
\providecommand{\BIBentrySTDinterwordspacing}{\spaceskip=0pt\relax}
\providecommand{\BIBentryALTinterwordstretchfactor}{4}
\providecommand{\BIBentryALTinterwordspacing}{\spaceskip=\fontdimen2\font plus
\BIBentryALTinterwordstretchfactor\fontdimen3\font minus
  \fontdimen4\font\relax}
\providecommand{\BIBforeignlanguage}[2]{{%
\expandafter\ifx\csname l@#1\endcsname\relax
\typeout{** WARNING: IEEEtran.bst: No hyphenation pattern has been}%
\typeout{** loaded for the language `#1'. Using the pattern for}%
\typeout{** the default language instead.}%
\else
\language=\csname l@#1\endcsname
\fi
#2}}
\providecommand{\BIBdecl}{\relax}
\BIBdecl

\bibitem{crash_report}
N.~C. for Statistics and M.~Analysis.~(2022, ``{Pedestrians: 2020 data (Traffic
  Safety Facts. Report No. DOT HS 813 310). National Highway Traffic Safety
  Administration.}''
  \url{https://crashstats.nhtsa.dot.gov/Api/Public/ViewPublication/813310},
  2022, [Online; accessed 26-Jan-2023].

\bibitem{urban_survey}
D.~Ridel, E.~Rehder, M.~Lauer, C.~Stiller, and D.~Wolf, ``A literature review
  on the prediction of pedestrian behavior in urban scenarios,'' in \emph{2018
  21st International Conference on Intelligent Transportation Systems (ITSC)},
  2018, pp. 3105--3112.

\bibitem{urban_survey2}
A.~Rasouli and J.~K. Tsotsos, ``Autonomous vehicles that interact with
  pedestrians: A survey of theory and practice,'' \emph{IEEE Transactions on
  Intelligent Transportation Systems}, vol.~21, no.~3, pp. 900--918, 2020.

\bibitem{deepsort}
N.~Wojke, A.~Bewley, and D.~Paulus, ``Simple online and realtime tracking with
  a deep association metric,'' in \emph{2017 IEEE International Conference on
  Image Processing (ICIP)}, 2017, pp. 3645--3649.

\bibitem{wang2018repulsion}
X.~Wang, T.~Xiao, Y.~Jiang, S.~Shao, J.~Sun, and C.~Shen, ``Repulsion loss:
  Detecting pedestrians in a crowd,'' in \emph{Proceedings of the IEEE
  conference on computer vision and pattern recognition}, 2018, pp. 7774--7783.

\bibitem{zhang2018occlusion}
S.~Zhang, L.~Wen, X.~Bian, Z.~Lei, and S.~Z. Li, ``Occlusion-aware r-cnn:
  detecting pedestrians in a crowd,'' in \emph{Proceedings of the European
  conference on computer vision (ECCV)}, 2018, pp. 637--653.

\bibitem{xu2018encoding}
Y.~Xu, Z.~Piao, and S.~Gao, ``Encoding crowd interaction with deep neural
  network for pedestrian trajectory prediction,'' in \emph{Proceedings of the
  IEEE conference on computer vision and pattern recognition}, 2018, pp.
  5275--5284.

\bibitem{korbmacher2022review}
R.~Korbmacher and A.~Tordeux, ``Review of pedestrian trajectory prediction
  methods: Comparing deep learning and knowledge-based approaches,'' \emph{IEEE
  Transactions on Intelligent Transportation Systems}, 2022.

\bibitem{shi2022social}
L.~Shi, L.~Wang, C.~Long, S.~Zhou, F.~Zheng, N.~Zheng, and G.~Hua, ``Social
  interpretable tree for pedestrian trajectory prediction,'' in
  \emph{Proceedings of the AAAI Conference on Artificial Intelligence},
  vol.~36, no.~2, 2022, pp. 2235--2243.

\bibitem{ferguson2015real}
S.~Ferguson, B.~Luders, R.~C. Grande, and J.~P. How, ``Real-time predictive
  modeling and robust avoidance of pedestrians with uncertain, changing
  intentions,'' in \emph{Algorithmic Foundations of Robotics XI: Selected
  Contributions of the Eleventh International Workshop on the Algorithmic
  Foundations of Robotics}.\hskip 1em plus 0.5em minus 0.4em\relax Springer,
  2015, pp. 161--177.

\bibitem{schneider2013pedestrian}
N.~Schneider and D.~M. Gavrila, ``Pedestrian path prediction with recursive
  bayesian filters: A comparative study,'' in \emph{Pattern Recognition: 35th
  German Conference, GCPR 2013, Saarbr{\"u}cken, Germany, September 3-6, 2013.
  Proceedings 35}.\hskip 1em plus 0.5em minus 0.4em\relax Springer, 2013, pp.
  174--183.

\bibitem{bouhsain2020pedestrian_LSTMPaper}
S.~A. Bouhsain, S.~Saadatnejad, and A.~Alahi, ``Pedestrian intention
  prediction: A multi-task perspective,'' \emph{arXiv preprint
  arXiv:2010.10270}, 2020.

\bibitem{HMM}
R.~Quintero, I.~Parra, J.~Lorenzo, D.~Fernández-Llorca, and M.~A. Sotelo,
  ``Pedestrian intention recognition by means of a hidden markov model and body
  language,'' in \emph{2017 IEEE 20th International Conference on Intelligent
  Transportation Systems (ITSC)}, 2017, pp. 1--7.

\bibitem{pedestrian_pose}
S.~Zhang, M.~Abdel-Aty, Y.~Wu, and O.~Zheng, ``Pedestrian crossing intention
  prediction at red-light using pose estimation,'' \emph{IEEE Transactions on
  Intelligent Transportation Systems}, vol.~23, no.~3, pp. 2331--2339, 2021.

\bibitem{spatiotemporal}
B.~Liu, E.~Adeli, Z.~Cao, K.-H. Lee, A.~Shenoi, A.~Gaidon, and J.~C. Niebles,
  ``Spatiotemporal relationship reasoning for pedestrian intent prediction,''
  \emph{IEEE Robotics and Automation Letters}, vol.~5, no.~2, pp. 3485--3492,
  2020.

\bibitem{gujjar2019classifying}
P.~Gujjar and R.~Vaughan, ``Classifying pedestrian actions in advance using
  predicted video of urban driving scenes,'' in \emph{2019 International
  Conference on Robotics and Automation (ICRA)}.\hskip 1em plus 0.5em minus
  0.4em\relax IEEE, 2019, pp. 2097--2103.

\bibitem{JAAD_cite1}
A.~Rasouli, I.~Kotseruba, and J.~K. Tsotsos, ``Are they going to cross? a
  benchmark dataset and baseline for pedestrian crosswalk behavior,'' in
  \emph{ICCVW}, 2017, pp. 206--213.

\bibitem{JAAD_cite2}
------, ``It’s not all about size: On the role of data properties in
  pedestrian detection,'' in \emph{ECCVW}, 2018.

\bibitem{PIE_dataset}
A.~Rasouli, I.~Kotseruba, T.~Kunic, and J.~K. Tsotsos, ``Pie: A large-scale
  dataset and models for pedestrian intention estimation and trajectory
  prediction,'' in \emph{Proceedings of the IEEE/CVF International Conference
  on Computer Vision (ICCV)}, October 2019.

\bibitem{psi_dataset}
T.~Chen, R.~Tian, Y.~Chen, J.~Domeyer, H.~Toyoda, R.~Sherony, T.~Jing, and
  Z.~Ding, ``Psi: A pedestrian behavior dataset for socially intelligent
  autonomous car,'' \emph{arXiv preprint arXiv:2112.02604}, 2021.

\bibitem{CARLA}
A.~Dosovitskiy, G.~Ros, F.~Codevilla, A.~Lopez, and V.~Koltun, ``{CARLA}: {An}
  open urban driving simulator,'' in \emph{Proceedings of the 1st Annual
  Conference on Robot Learning}, 2017, pp. 1--16.

\bibitem{yolov5}
\BIBentryALTinterwordspacing
G.~Jocher, ``Yolov5 by ultralytics,'' 5 2020. [Online]. Available:
  \url{https://github.com/ultralytics/yolov5}
\BIBentrySTDinterwordspacing

\bibitem{attention}
A.~Vaswani, N.~Shazeer, N.~Parmar, J.~Uszkoreit, L.~Jones, A.~N. Gomez,
  L.~Kaiser, and I.~Polosukhin, ``Attention is all you need,'' \emph{CoRR},
  vol. abs/1706.03762, 2017.

\bibitem{ROS}
M.~Quigley, K.~Conley, B.~Gerkey, J.~Faust, T.~Foote, J.~Leibs, R.~Wheeler,
  A.~Y. Ng \emph{et~al.}, ``Ros: an open-source robot operating system,'' in
  \emph{ICRA workshop on open source software}, vol.~3, no. 3.2.\hskip 1em plus
  0.5em minus 0.4em\relax Kobe, Japan, 2009, p.~5.

\bibitem{mscoco}
T.-Y. Lin, M.~Maire, S.~Belongie, J.~Hays, P.~Perona, D.~Ramanan,
  P.~Doll{\'a}r, and C.~L. Zitnick, ``Microsoft coco: Common objects in
  context,'' in \emph{Computer Vision--ECCV 2014: 13th European Conference,
  Zurich, Switzerland, September 6-12, 2014, Proceedings, Part V 13}.\hskip 1em
  plus 0.5em minus 0.4em\relax Springer, 2014, pp. 740--755.

\end{thebibliography}

\end{document}